\documentclass{article}

\usepackage{arxiv}
\renewcommand{\headeright}{Preprint}  %
\renewcommand{\undertitle}{Preprint}  %

\usepackage[utf8]{inputenc} %
\usepackage[T1]{fontenc}    %
\usepackage{times}
\usepackage{xspace}
\usepackage[dvipsnames]{xcolor}
\usepackage{graphicx}
\usepackage{amsmath}
\usepackage{amssymb}
\usepackage{amsfonts}
\usepackage{amsthm}
\usepackage{booktabs}
\usepackage{microtype}
\usepackage[numbers,sort&compress]{natbib}
\usepackage[format=plain,labelformat=simple,labelsep=period,font=small,compatibility=false]{caption}
\usepackage[font=footnotesize,skip=3pt,subrefformat=parens]{subcaption}
\usepackage[hyphens]{url}
\usepackage{array}
\usepackage{tabularx}
\usepackage{multirow}
\usepackage{algorithm}
\usepackage{algpseudocode}
\usepackage{verbatim}
\usepackage{tikz}
\usetikzlibrary{positioning, arrows.meta, calc, shapes.geometric, shapes.misc}

\newtheorem{proposition}{Proposition}

\definecolor{cvprblue}{rgb}{0.21,0.49,0.74}
\definecolor{ssOrange}{HTML}{D55E00}

\usepackage{xr-hyper}
\usepackage[pagebackref,breaklinks,colorlinks,allcolors=cvprblue]{hyperref}
\usepackage{doi}

\makeatletter
\DeclareRobustCommand\onedot{\futurelet\@let@token\@onedot}
\def\@onedot{\ifx\@let@token.\else.\null\fi\xspace}
\makeatother

\newcommand{\tablefont}{\fontsize{7}{8.5}\selectfont}

\newcommand{\sbvideo}{Social Signal Bench~Video\xspace}
\newcommand{\sbaudio}{Social Signal Bench~Audio\xspace}

\usepackage{chapterbib}

\title{Social Signal Bench: Evidence-Weighted Evaluation of\\Co-occurring Social Signals in Video and Audio}

\author{
	Siddharth Ravi$^{1,2}$ \quad
	Anders R.~Bargum$^{1}$ \quad
	Hugo Anselme$^{1,3,4}$ \quad
	Patrizio Cugia di Sant'Orsola$^{1}$ \\
	\bfseries
	Alberto Cardaci$^{1}$ \quad
	Flora Kincso Parragi$^{1}$ \quad
	Vincent Borowiec$^{1}$ \quad
	Line H Clemmensen$^{5}$ \\[0.8em]
	\normalfont\small
	$^{1}$Interhuman AI \quad
	$^{2}$Universidad de Alicante \quad
	$^{3}$École Polytechnique \\
	\normalfont\small
	$^{4}$Institut Supérieure d'Electronique de Paris \quad
	$^{5}$University of Copenhagen
}

\renewcommand{\shorttitle}{Social Signal Bench}

\hypersetup{
pdftitle={Social Signal Bench: Evidence-Weighted Evaluation of Co-occurring Social Signals in Video and Audio},
pdfauthor={Siddharth Ravi, Anders R. Bargum, Hugo Anselme, Patrizio Cugia di Sant'Orsola, Alberto Cardaci, Flora Kincso Parragi, Vincent Borowiec, Line H Clemmensen},
pdfkeywords={social signal recognition, affective computing, multimodal benchmark, audio-visual understanding, omni-modal models, multimodal large language models, nonverbal behavior, engagement recognition, post-training},
}

\begin{document}
\maketitle
\suppressfloats[t] %

\begin{abstract}
Recognizing social signals such as interest and hesitation requires interpreting context-dependent behaviors that may co-occur in conversation. Existing affective computing benchmarks rarely evaluate such co-occurring signals, and conventional multi-label F1 overlooks semantic relationships and annotator disagreement. We introduce \textbf{Social Signal Bench}, a 47-hour densely annotated benchmark for the evaluation of social signal recognition and engagement, comprising two datasets: \textbf{\sbvideo} (25 hours), covering ten social signals in audiovisual recordings, and \textbf{\sbaudio} (22 hours), covering six signals in audio-only recordings. Each dataset pairs a public training set of online conversational video, of which 10 hours are expert-annotated and the remainder pseudo-labeled, with a private expert-labeled test set spanning four conversational domains. We also propose \textbf{Evidence-Weighted F1 (EW-F1)}, which weights reference labels by annotator evidence and assigns graded credit to semantically related predictions through optimal per-clip matching. Controlled perturbations demonstrate that EW-F1 distinguishes semantically related errors from unrelated ones, unlike conventional F1. Under a shared evaluation protocol, 7B open models post-trained on Social Signal Bench outperform general-purpose frontier models on video and match them on audio.
\end{abstract}


\begin{figure}[t]
\centering
\scalebox{0.7778}{\input{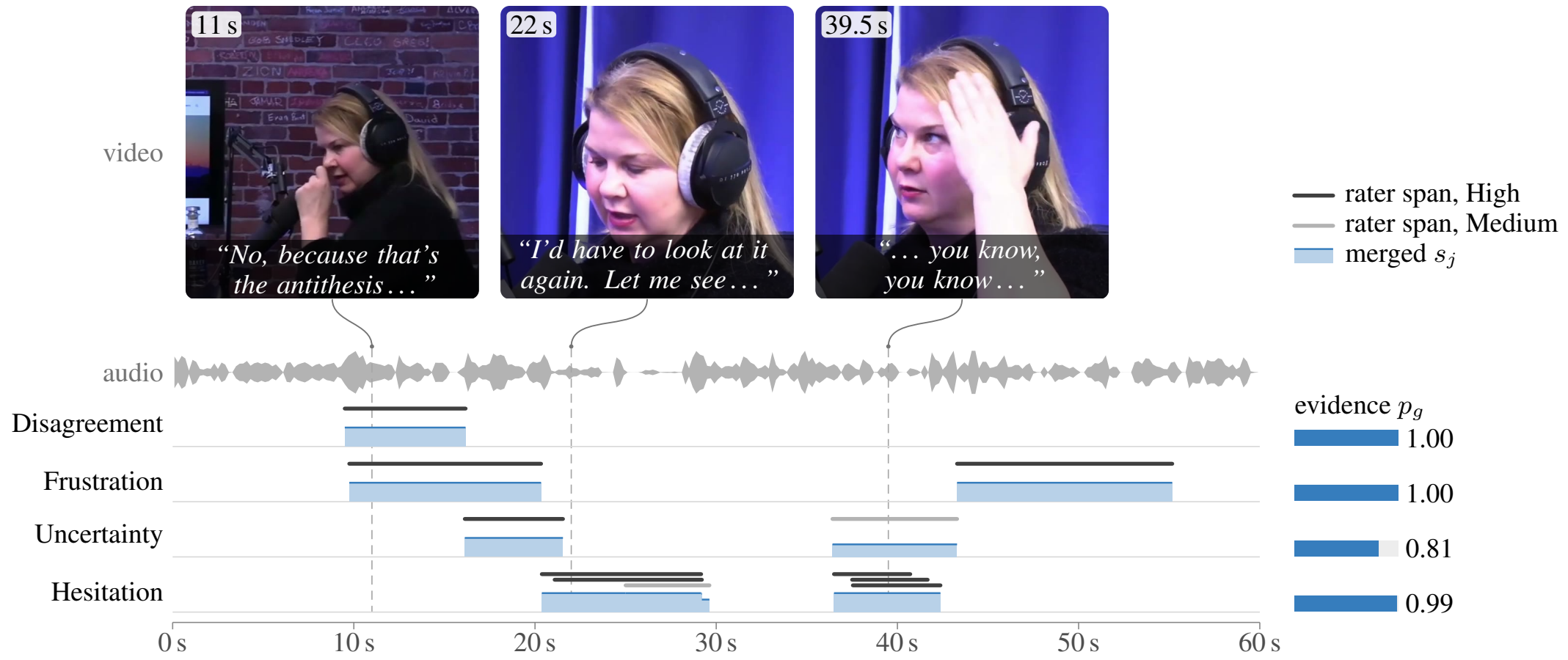}}
\caption{\textbf{Several social signals annotated on one minute of conversation.} Three trained raters mark when each signal occurs. Each line is one rater's span: dark lines mean the rater was highly confident, light lines moderately confident. The blue band shows how strong the combined evidence is at each moment, and the bars on the right give each signal's overall evidence weight (Sec.~\ref{sec:annotation}). Evidence weights keep all four signals, and one highly confident rater is enough for full weight. Engagement is labeled separately and not shown.}
\label{fig:teaser}
\end{figure}

\section{Introduction}
\label{sec:introduction}
Social signals are states such as hesitation or skepticism that observers infer from nonverbal and paralinguistic cues, such as pauses and changes of tone~\citep{vinciarelli2012bridging,poggi2012social}. Reading these signals from video and audio is a long-standing goal of affective computing and social signal processing~\citep{picard1997affective,zeng2009survey,vinciarelli2009ssp}, and omni-modal foundation models now take both streams as input~\citep{xu2025qwen25omni}. Listeners form several such judgments about the same moment at once~\citep{vinciarelli2009ssp,ambady1992thin}. Video and omni-modal language models describe what is visible and audible well and lead general audio-visual benchmarks~\citep{zhang2023videollama,cheng2024videollama2,gemini2024gemini15,hurst2024gpt4o,fu2025videomme,li2025omnibench,hong2026worldsense}, but reading social signals also requires reasoning about interpersonal dynamics, where current models remain limited~\citep{mathur2025socialgenome,jin2024mmtom}.

Measuring this ability is difficult in two ways. On the signal side, cues and signals map many-to-many: one cue can support several interpretations, and one signal can be expressed through different cues~\citep{Brooks2024DeepLearning, Goel2024FaceContext}. Interpretation also depends on the interactional setting and on preceding and, when available, subsequent behavior~\citep{Goel2024FaceContext, Zhang2025ECERC}, and culture and speaker style change how behavior is expressed and read~\citep{Li2025CulturesEmotion, EnsenbergDiamant2025IndividualDifferences, AbdurRakib2026Personality}. Signals co-occur, so the task is multi-label~\citep{Vidrascu2005AnnotationAD}. On the observer side, trained raters can reasonably disagree, and such disagreement often reflects genuine ambiguity rather than annotation noise~\citep{Shinoda2024AnnotatorBias, Fleisig2024Perspectivist, larsen2026elusive}.

Multi-label recognition is commonly scored with variants of the F1 score~\citep{zhang2014review}. Multi-label F1 treats all classes as orthogonal, and predicting related or close signals costs as much as predicting unrelated or distant ones. It also treats the reference as certain. A label that every rater marked with high confidence and a label that only one rater inferred carry the same recall mass. Under the standard F1 scoring, a model that omits a label it is unsure of can therefore score higher than a model that predicts a closely related one (Figure~\ref{fig:matching}). Exact F1 thus favors omission over a near miss, and F1-optimal decision thresholds are known to favor abstention on rare labels~\citep{lipton2014optimal}.

\begin{figure*}[t]
\centering
\colorlet{ssEvidence}{cvprblue}%
\begin{tikzpicture}[
    font=\scriptsize,
    card/.style={draw=black!35, fill=white, rounded corners=2pt, align=center, text width=1.85cm,
                 minimum height=1.0cm, inner sep=2pt},
    expert/.style={card, draw=cvprblue!70, line width=0.6pt},
    teacher/.style={card, draw=black!35},
    evid/.style={card, draw=ssEvidence, line width=0.6pt},
    split/.style={card, fill=black!4, draw=black!50, line width=0.6pt},
    tax/.style={card, fill=cvprblue!10, draw=cvprblue!70, text width=2.6cm, minimum height=0.8cm},
    metric/.style={card, fill=black, text=white, draw=black},
    flow/.style={-{Stealth[length=1.7mm]}, black!70, line width=0.7pt},
    soft/.style={flow, densely dashed, black!45},
    uses/.style={-{Stealth[length=1.6mm]}, cvprblue!80, densely dashed, line width=0.6pt},
    phase/.style={fill=black!3, rounded corners=4pt},
    phaselab/.style={anchor=north west, font=\scriptsize\bfseries, text=black!55, inner sep=1pt},
    elab/.style={inner sep=1pt, fill=white, font=\scriptsize},
]
  \def\yA{0}  \def\yB{-1.65}
  \def\st#1{{\fontsize{8}{9.5}\selectfont\bfseries #1}}   %
  \def\sd#1{{\color{black!60}#1}}           %

  \fill[phase] (0,1.0) rectangle (11.17,-2.3);
  \fill[phase] (11.32,1.0) rectangle (13.78,-2.3);
  \fill[phase] (13.93,1.0) rectangle (16.3,-2.3);
  \node[phaselab] at (0.08,0.95)  {1\;\;Build the reference};
  \node[phaselab] at (11.4,0.95) {2\;\;Release};
  \node[phaselab] at (14.01,0.95) {3\;\;Evaluate};

  \node[expert] (col) at (1.1,\yA)  {\st{Collect}\\[0.5pt]\sd{recordings from\\four domains}};
  \node[expert] (ann) at (3.35,\yA)  {\st{Annotate}\\[0.5pt]\sd{${\geq}3$ trained\\raters per clip}};
  \node[expert] (rev) at (5.6,\yA)  {\st{Review}\\[0.5pt]\sd{cluster-severe\\votes only}};
  \node[evid]   (mrg) at (7.85,\yA)  {\st{Merge}\\[0.5pt]\sd{slice evidence $s_j$\\$\rightarrow$ weight $p_g$}};
  \node[expert] (seg) at (10.1,\yA) {\st{Segment}\\[0.5pt]\sd{5--20\,s clips, split\\by recording}};
  \foreach \a/\b in {col/ann, ann/rev, rev/mrg, mrg/seg} \draw[flow] (\a) -- (\b);

  \node[teacher] (tch) at (3.35,\yB) {\st{Train teacher}\\[0.5pt]\sd{on proprietary\\annotated data}};
  \node[teacher] (psd) at (5.6,\yB) {\st{Pseudo-label}\\[0.5pt]\sd{additional\\online video}};
  \draw[flow] (tch) -- (psd);
  \draw[soft] (ann) -- node[elab, right=1pt, fill=black!3, text=black!55] {same protocol} (tch);

  \node[split] (test)  at (12.55,\yA) {\st{Private test}\\[0.5pt]\sd{four domains,\\expert labels}};
  \node[split] (train) at (12.55,\yB) {\st{Public train}\\[0.5pt]\sd{online video,\\expert + pseudo}};
  \draw[flow] (seg) -- (test);
  \draw[flow] (seg.south) -- ++(0,-0.3) -| ([xshift=-0.35cm]train.north);
  \draw[flow] (psd) -- (train);

  \node[metric] (ew)  at (15.12,\yA) {\st{EW-F1}\\[0.5pt]{\color{white!80!black}corpus-level,\\hosted}};
  \node[card]   (mod) at (15.12,\yB) {\st{Model}\\[0.5pt]\sd{zero-shot or\\fine-tuned}};
  \draw[soft] (train) -- (mod);
  \draw[flow] (mod) -- (ew);
  \draw[flow, ssEvidence] (test) -- node[elab, above=1pt, fill=black!3, text=ssEvidence] {$p_g$} (ew);

  \node[tax] (tx) at (10.1,1.75) {\st{Taxonomy}\\\sd{12 signals, 7 clusters}};
  \draw[uses] (tx.west) -| node[elab, pos=0.25, above=0pt, fill=none] {disagreement severity} (rev.north);
  \draw[uses] (tx.east) -| node[elab, pos=0.25, above=0pt, fill=none] {graded credit $c(q,g)$} ([xshift=0.5cm]ew.north);  %
\end{tikzpicture}
\caption{\textbf{Social Signal Bench construction and evaluation.} Each card is a pipeline step, and its grey text describes that step. Expert annotations (top lane) pass through taxonomy-guided review and an evidence merge that gives each reference signal a weight $p_g$. A teacher trained on proprietary data annotated with the same protocol pseudo-labels additional online video for the public training set (bottom lane). The private test sets hold expert labels only. EW-F1 uses the same taxonomy for graded credit and weights each reference signal by $p_g$.}
\label{fig:pipeline}
\end{figure*}
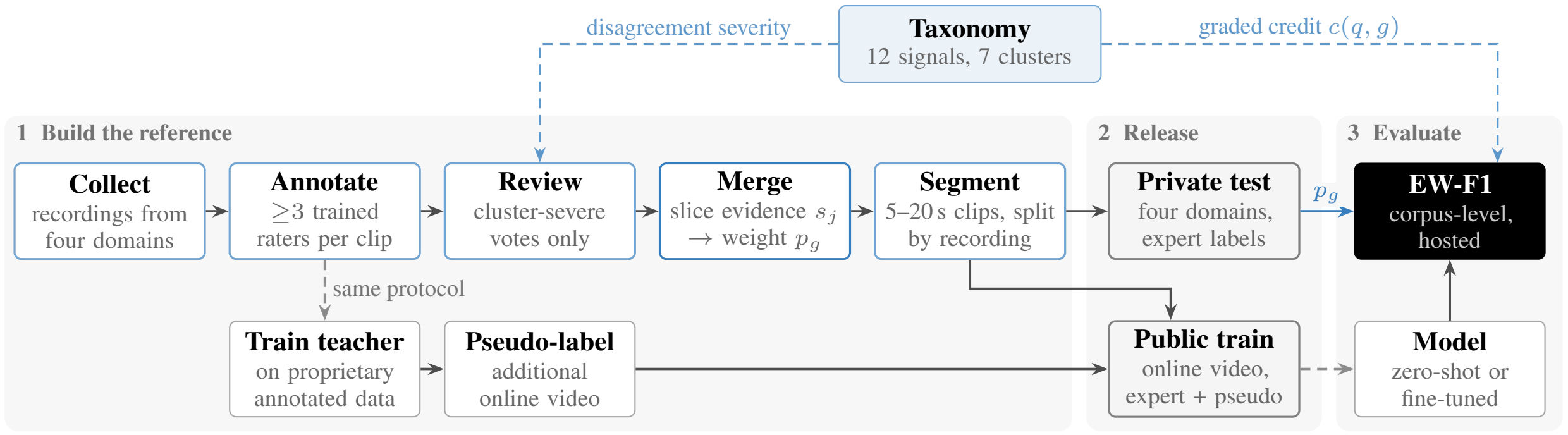

Social Signal Bench ties the metric to how the reference was annotated (Figure~\ref{fig:pipeline}). The taxonomy that guides annotation and disagreement review also sets partial credit for related predictions, and accumulated annotator evidence sets the weight of each reference label. Our contributions are:
\begin{itemize}
     \item \textbf{A social-signal benchmark} comprising \textbf{\sbvideo} and \textbf{\sbaudio}. \sbvideo covers ten co-occurring social signals (agreement, confidence, interest, frustration, uncertainty, hesitation, stress, confusion, skepticism, and disagreement; Figure~\ref{fig:taxonomy}), and \sbaudio covers the first six. Both datasets include public training sets with 10\,h of expert annotations supplemented by teacher-generated pseudo-labels, and private expert-labeled test sets spanning four conversational domains (2{,}372 video and 1{,}094 audio clips). Expert-annotated clips are independently assessed by at least three trained raters, capturing temporal spans, intensity, confidence, and engagement.

    \item \textbf{Evidence-Weighted F1 (EW-F1)}, a corpus-level metric that combines taxonomy-graded credit with annotation evidence weights, validated with controlled perturbations and parameter-sensitivity analyses.

    \item \textbf{An evaluation protocol} with reference results for proprietary and open-weight frontier models, task-specific post-trained models, and human raters under a shared harness.

    \item \textbf{A publicly accessible benchmark}, of which this paper describes the first release, providing both Social Signal Bench datasets for model development, hosted evaluation on private expert-labeled test sets, a public leaderboard, and open-source evaluation code.
\end{itemize}
The benchmark and the post-trained models, including the training annotations, taxonomy, EW-F1 scorer, and evaluation schemas, will be released at \url{https://huggingface.co/datasets/Interhuman/social-signal-bench}. Future releases may expand the data, with evaluation settings fixed within each version.

\section{Related Work}
\label{sec:related-work}

\paragraph{Affective and social-interaction corpora.}
Emotion corpora such as IEMOCAP~\citep{busso2008iemocap}, MELD~\citep{poria2019meld}, CMU-MOSEI~\citep{zadeh2018cmumosei}, and M3ED~\citep{zhao2022m3ed} target categorical emotion, sentiment, or emotion intensity at utterance scale. RECOLA~\citep{ringeval2013recola} provides continuous valence and arousal traces during remote collaboration, and NoXi~\citep{cafaro2017noxi} annotates engagement and interaction descriptors in mediated novice-expert sessions. DAiSEE~\citep{gupta2016daisee} labels four learning-related affective states in e-learning video. Social-IQ~\citep{zadeh2019socialiq} probes social reasoning with video question answering, and the MER challenges~\citep{lian2024mer,lian2025mer2025} add open-vocabulary and descriptive emotion tracks. Social Genome~\citep{mathur2025socialgenome} and MMToM-QA~\citep{jin2024mmtom} evaluate grounded social and mental-state reasoning in multimodal models through question answering. General video and omni-modal benchmarks, such as MVBench~\citep{li2024mvbench}, Video-MME~\citep{fu2025videomme}, OmniBench~\citep{li2025omnibench}, and WorldSense~\citep{hong2026worldsense}, test temporal, perceptual, and audio-visual understanding with multiple-choice questions. Dense video datasets such as AVA~\citep{gu2018ava} annotate co-occurring atomic actions in space and time, but not perceived social signals. EmoBench-M~\citep{hu2026emobenchm} aggregates many of these sources into a benchmark for multimodal language models. MERR~\citep{cheng2024emotionllama} and MER-Caption~\citep{lian2025affectgpt} scale descriptive emotion data with labels and rationales written by language models, with human verification done on a subset. 
Unlike these corpora, Social Signal Bench targets co-occurring social signals with dense, independent annotations from trained raters, and keeps temporal evidence, confidence, and disagreement instead of a single consensus label; its private test sets span four conversational domains at natural signal prevalence.

\paragraph{Learning from disagreement.}
Redundant labeling reduces individual noise, and annotator-reliability models estimate a latent truth from noisy raters~\citep{snow2008cheap,sheng2008get, dawid1979maximum,raykar2010learning}. Prior work argues that disagreement on subjective tasks should be preserved and evaluated with soft references~\citep{uma2021disagreement,plank2022problem}. Training on full human label distributions is also shown to improve robustness in image classification, and jury learning models individual annotators instead of a single consensus~\citep{peterson2019human, gordon2022jury}. Social Signal Bench follows this view with regard to both the reference and the metric. It reviews only disagreements that the taxonomy marks as implausible, and it carries the remaining evidence into the score.

\paragraph{Graded and evidence-aware evaluation.}
Most multi-label metrics treat every wrong label as equally wrong. Hierarchical classification measures give partial credit by distance in a class hierarchy~\citep{kosmopoulos2015evaluation}, and work in visual recognition trades accuracy against specificity or scores the severity of a mistake by its distance in the hierarchy~\citep{deng2012hedging,bertinetto2020making}. CrowdTruth instead keeps the spread of crowd annotations and weights each label by how strongly the crowd supports it~\citep{aroyo2015truth}. EW-F1 combines the two ideas for set-valued predictions: the taxonomy sets partial credit, and accumulated rater evidence sets the weight of each reference label.

\paragraph{Holistic versus cue-level annotation.}
Thin-slice research shows that brief behavioral observations support meaningful interpersonal judgments and that holistic judgments can predict better than ratings of individual micro-behaviors~\citep{ambady1992thin, ambady1993half}. Cues do not map deterministically onto social signals, may differ between modalities, and are read by integrating several cues with context, with substantial variability across situations and observers~\citep{barrett2019emotional, Goel2024FaceContext}. Social Signal Bench therefore asks trained raters for integrated judgments of perceived social signals, guided by operational definitions and cue examples rather than cue checklists. This captures context-dependent interpretation but introduces observer variability~\citep{Shinoda2024AnnotatorBias, Tavernor2024Annotators}, which we address with calibration, independent annotations with confidence ratings, and EW-F1.

\begin{table*}[t]
\centering
\caption{Comparison with related multimodal corpora. Units differ between rows (turns, utterances, sentences, sessions, clips), so the table compares task and annotation design, not scale or difficulty.}
\label{tab:related-corpora}
\tablefont
\setlength{\tabcolsep}{2.5pt}
\begin{tabularx}{\textwidth}{@{}>{\raggedright\arraybackslash}p{41pt}>{\raggedright\arraybackslash}p{140pt}>{\raggedright\arraybackslash}p{94pt}>{\raggedright\arraybackslash}p{45pt}>{\raggedright\arraybackslash}X@{}}
\toprule
\textbf{Dataset} & \textbf{Target} & \textbf{Unit and scale} & \textbf{Modalities} & \textbf{Reference and disagreement} \\
\midrule
IEMOCAP & Categorical and dimensional emotion~\citep{busso2008iemocap} & 10{,}039 turns; $\sim$12\,h; 10 actors & AV, text, mocap & Multiple raters; categorical agreement reported \\
MELD & Emotion, sentiment~\citep{poria2019meld} & 13{,}708 utterances; TV dialogue & AV, text & Three raters; majority label; Fleiss' $\kappa=0.43$ \\
CMU-MOSEI & Sentiment, six emotion intensities~\citep{zadeh2018cmumosei} & 23{,}453 sentences; 1{,}000 speakers & AV, text & Three crowd raters; averaged ratings \\
M3ED & Seven emotions, blends~\citep{zhao2022m3ed} & 24{,}449 utterances; 56 TV series & AV, text & Three raters; majority; Fleiss' $\kappa=0.59$ \\
RECOLA & Valence, arousal, social behavior~\citep{ringeval2013recola} & 3.8\,h annotated; 46 participants & AV, ECG, EDA & Six raters; continuous traces \\
NoXi & Engagement, interaction descriptors~\citep{cafaro2017noxi} & 84 sessions; 25\,h & AV, depth & Continuous and discrete annotations \\
DAiSEE & Boredom, confusion, engagement, frustration~\citep{gupta2016daisee} & 9{,}068 snippets; 112 users & Video & Crowd labels; psychologist reference subset \\
Social-IQ & Social reasoning QA~\citep{zadeh2019socialiq} & 1{,}250 videos; 7{,}500 questions & AV, text & Correct/incorrect answers \\
\midrule
\textbf{\sbvideo}
    & Ten co-occurring social signals + engagement status
    & 10{,}173 clips; 25\,h; 857 sources; four domains
    & AV, transcript context
    & 15.6\,h expert-annotated (10\,h training + test);
      remaining 9.5\,h teacher-pseudo-labeled \\
\addlinespace[3pt]
\textbf{\sbaudio}
    & Six co-occurring social signals + engagement status
    & 8{,}392 clips; 22\,h; 806 sources; four domains
    & Audio, transcript context
    & 12.4\,h expert-annotated (10\,h training + test);
      remaining 9.6\,h teacher-pseudo-labeled \\
\bottomrule
\end{tabularx}
\end{table*}

\section{The Social Signal Bench Benchmark}
\label{sec:benchmark}

Social Signal Bench comprises two datasets: 25 hours of audiovisual data (10{,}173 clips) and 22 hours of audio-only data (8{,}392 clips). Each dataset consists of a public training set of online conversational video and a private, expert-labeled test set spanning four conversational domains (Table~\ref{tab:release}). Each public training set contains 10 hours of expert-annotated clips; the remaining training data are pseudo-labeled by teacher models trained on a larger proprietary social-signal dataset from all four domains. Both follow a shared multi-stage pipeline depicted in Figure~\ref{fig:pipeline}.

Each item in Social Signal Bench is a short clip $x_i$, either video with audio or audio only. Given a clip, a system returns which of the taxonomy's perceived social signals, such as \emph{hesitation} or \emph{skepticism} (ten for video, six for audio; Section~\ref{sec:clusters}), are present, and a three-class engagement status (\texttt{disengaged}, \texttt{neutral}, \texttt{engaged}). An empty signal set is a valid answer: many clips of natural conversation show no social signal above the labeling threshold. Signals are scored with EW-F1 (Section~\ref{sec:ewf1}), and the engagement status independently with three-class macro-F1, on which a constant majority answer scores low (Supplement~\ref{app:engagement}). We evaluate clips under the natural class prevalence observed in the annotated data. Balancing the set by removing negative examples would change this prevalence and was, in our diagnostic, shown to inflate aggregate metrics (Supplement~\ref{app:construction}).

\subsection{Taxonomy}
\label{sec:clusters}
The taxonomy has twelve perceived social signals (Figure~\ref{fig:taxonomy}), describing what an observer perceives~\citep{vinciarelli2009ssp, mehu2012conceptual}. The signals were selected for their relevance to conversational feedback, initially in sales training, and each is defined from standard psychological sources~\citep{apa_dictionary}. Ten signals form the multi-label signal head. Engagement and disengagement form a separate three-class engagement head: we treat them as the two ends of one continuum of attention to the interaction and the other ten signals as departures from that baseline. Raters annotate the two ends, and time marked as neither counts as neutral. Signals that share observable cues and sit close together on affective and interpersonal dimensions, such as alignment versus opposition and approach versus withdrawal, are grouped into clusters~\citep{russell1980circumplex, mehu2014multimodal}. A surjection $\mathrm{Cluster}:\mathcal{L}\to\mathcal{C}$ maps every signal to one of seven clusters. A symmetric, irreflexive relation $\mathrm{Adj}\subseteq\mathcal{C}\times\mathcal{C}$ marks which cross-cluster confusions are plausible. \{Agreement, Confidence, Engagement \& Interest\} form one adjacency triangle, \{Skepticism \& Disagreement, Stress \& Frustration, Uncertainty \& Hesitation\} form another, and Disengagement is adjacent to none. In our annotation guide, adjacent clusters share more observable cues than non-adjacent ones; Uncertainty \& Hesitation and Skepticism \& Disagreement share six.

\begin{figure}[t]
\centering
\begin{minipage}[b]{0.48\textwidth}\centering
\resizebox{\linewidth}{!}{\begin{tikzpicture}[
    font=\scriptsize,
    group/.style={fill=black!4, rounded corners=4pt},
    clbox/.style={draw=cvprblue!60, fill=cvprblue!14, rounded corners=3pt, minimum width=1.6cm},
    sig/.style={fill=white, draw=cvprblue!45, rounded corners=1.5pt, minimum width=1.4cm,
                minimum height=0.3cm, inner sep=0.5pt},
    vonly/.style={sig, densely dashed, draw=cvprblue!70},
    head/.style={sig, draw=black!30, text=black!55, font=\scriptsize\itshape},
    adj/.style={cvprblue!70, line width=1.2pt},
    w/.style={fill=black!4, text=cvprblue!90!black, inner sep=0.8pt},
]
  \fill[group] (0,0.3) rectangle (3.95,3.5);
  \fill[group] (4.15,0.3) rectangle (8.1,3.5);

  \node[clbox, minimum height=0.5cm]  (agr)  at (1.975,2.95) {};
  \node[clbox, minimum height=0.5cm]  (conf) at (0.9,1.05) {};
  \node[clbox, minimum height=0.88cm] (eng)  at (3.05,1.05) {};
  \node[clbox, minimum height=0.88cm] (skep) at (6.125,2.95) {};
  \node[clbox, minimum height=0.88cm] (str)  at (5.05,1.05) {};
  \node[clbox, minimum height=1.26cm] (unc)  at (7.2,1.05) {};

  \foreach \a/\b in {agr/conf, agr/eng, skep/str, skep/unc}
    \draw[adj] (\a) -- node[w] {0.33} (\b);
  \foreach \a/\b in {conf/eng, str/unc}
    \draw[adj] (\a) -- node[w, above=1pt, fill=none] {0.33} (\b);

  \node[sig]   at (1.975,2.95) {Agreement};
  \node[sig]   at (0.9,1.05)  {Confidence};
  \node[sig]   at (3.05,1.24)   {Interest};
  \node[head]  at (3.05,0.86)   {Engagement};
  \node[vonly] at (6.125,3.14) {Disagreement};
  \node[vonly] at (6.125,2.76) {Skepticism};
  \node[sig]   at (5.05,1.24)   {Frustration};
  \node[vonly] at (5.05,0.86)   {Stress};
  \node[sig]   at (7.2,1.43)  {Uncertainty};
  \node[sig]   at (7.2,1.05)  {Hesitation};
  \node[vonly] at (7.2,0.67)  {Confusion};

  \node[clbox, draw=black!30, fill=black!4, minimum width=1.95cm, minimum height=0.5cm] at (4.05,-0.1) {};
  \node[head, minimum width=1.75cm] at (4.05,-0.1) {Disengagement};

  \def\ly{-0.72}
  \node[anchor=west, inner sep=0pt] at (0,\ly) {credit:\ same signal 1};
  \node[clbox, minimum width=0.4cm, minimum height=0.26cm] at (2.45,\ly) {};
  \node[anchor=west, inner sep=0pt] at (2.72,\ly) {same box 0.66};
  \draw[adj] (4.35,\ly) -- ++(0.4,0);
  \node[anchor=west, inner sep=0pt] at (4.85,\ly) {joined 0.33};
  \node[anchor=west, inner sep=0pt] at (6.5,\ly) {otherwise 0};
  \def\ly{-1.08}
  \node[vonly, minimum width=0.4cm, minimum height=0.24cm] at (0.2,\ly) {};
  \node[anchor=west, inner sep=0pt] at (0.48,\ly) {video only};
  \node[head, minimum width=0.4cm, minimum height=0.24cm] at (2.45,\ly) {};
  \node[anchor=west, inner sep=0pt, text=black!55] at (2.72,\ly) {engagement head, not scored by EW-F1};
\end{tikzpicture}}
\end{minipage}\hfill
\begin{minipage}[b]{0.48\textwidth}\centering
\resizebox{\linewidth}{!}{\begin{tikzpicture}[x=0.085cm, y=0.41cm, font=\scriptsize]
  \foreach \g in {0,20,40,60}
    \draw[gray!25] (\g,0.6) -- (\g,-9.6) node[below, text=black] {\g\%};
  \foreach \i/\name/\v in {0/Confidence/41.9, 1/Uncertainty/13.4, 2/Hesitation/10.2,
      3/Interest/8.0, 4/Frustration/7.1, 5/Agreement/4.1, 6/Skepticism/4.3,
      7/Stress/2.8, 8/Disagreement/2.4, 9/Confusion/2.0} {
    \node[left] at (0,-\i) {\name};
    \fill[cvprblue] (0,-\i+0.03) rectangle (\v,-\i+0.40);
    \node[right, inner sep=1pt, font=\tiny] at (\v,-\i+0.22) {\v};
  }
  \foreach \i/\a in {0/68.0, 1/14.7, 2/17.3, 3/9.2, 4/10.9, 5/3.5} {
    \fill[cvprblue!40] (0,-\i-0.40) rectangle (\a,-\i-0.03);
    \node[right, inner sep=1pt, font=\tiny] at (\a,-\i-0.22) {\a};
  }
  \foreach \i in {6,7,8,9}
    \node[right, inner sep=1pt, font=\tiny\itshape, text=gray] at (0,-\i-0.22) {video only};
  \fill[cvprblue] (40,-6.6) rectangle (43.5,-6.25);
  \node[right] at (43.5,-6.42) {Video};
  \fill[cvprblue!40] (40,-7.5) rectangle (43.5,-7.15);
  \node[right] at (43.5,-7.32) {Audio};
\end{tikzpicture}}
\end{minipage}

\medskip
\begin{minipage}[t]{0.48\textwidth}
\caption{The twelve-signal taxonomy. Boxes are clusters and lines join adjacent clusters, which sets the graded credit of \mbox{EW-F1}. Engagement and Disengagement form a separate head; dashed signals are video-only (Supplement Table~\ref{tab:signal-taxonomy}).}
\label{fig:taxonomy}
\end{minipage}\hfill
\begin{minipage}[t]{0.48\textwidth}
\caption{Share of clips in the public training sets with each signal at any probability (low, medium, or high; expert and pseudo-labels combined). No signal: 29.1\% of video clips and 15.2\% of audio clips.}
\label{fig:signal-dist}
\end{minipage}
\end{figure}

\subsection{Data sources and composition}
\label{sec:corpus}
Social Signal Bench draws on four conversational domains. Two are collections of naturalistic online conversational videos. The third is recorded sales calls. The fourth is long-form dyadic conversations (about 40 minutes per session), recorded over video conferencing with isolated audio and video tracks. These sessions follow a structured elicitation protocol (Supplement~\ref{app:construction}) that guides turns without scripting utterances, so disagreement, skepticism, hesitation, and agreement arise and resolve during the session.

Confidence is the most frequent signal detected in online video and sales calls in our dataset, whereas uncertainty leads in the dyadic conversations. Sales calls and dyadic conversations come from few, long recordings: of the 85 recordings in the \sbvideo test set, 75 are online video, 6 sales calls, and 4 dyadic conversations. The two modalities are annotated independently, so their label distributions differ (Table~\ref{tab:release}, Figure~\ref{fig:signal-dist}). Both keep naturally occurring empty-label clips instead of balancing signal prevalence, so models must recognize when no signal is sufficiently supported.

\paragraph{Public training sets and private test sets.}
Because the sales and dyadic recordings cannot be redistributed, the public training sets contain only online video, and every released clip comes from a CC~BY~3.0 licensed YouTube video (Table~\ref{tab:release}). Their expert-annotated clips are selected from online-video recordings outside the test partitions; the pseudo-labeled clips come from additional online videos, labeled by the video and audio teacher models, whose training data exclude every test recording. On held-out expert-annotated clips, the teacher models score 54.8 EW-F1 [50.4, 61.0] on video and 65.4 [62.2, 69.1] on audio (Supplement~\ref{sec:supp-post-training}). The audio score is skewed by the concentration of \textit{confidence} labels in the evaluation set. The test partitions stay private to not enter web-scale pretraining data~\citep{sainz2023nlp}, cover all four domains, and share no recording, channel, or participant with the public data. %

\begin{table}[t]
\centering
\caption{Composition of Social Signal Bench. The training sets are public; the test sets are private and span all four domains. \emph{Empty}: clips with no annotated signal at any probability.}
\label{tab:release}
\tablefont
\setlength{\tabcolsep}{3pt}
\begin{tabular*}{\columnwidth}{@{\extracolsep{\fill}}llrrrr@{}}
\toprule
\textbf{Split} & \textbf{Labels} & \textbf{Clips} & \textbf{Hours} & \textbf{Sources} & \textbf{Empty} \\
\midrule
\multicolumn{6}{@{}l}{\textbf{\sbvideo}} \\
\midrule
Train (public) & Expert & 4{,}321 & 10.0 & 319 & 37.2\% \\
Train (public) & Pseudo & 3{,}480 & 9.5 & 453 & 19.0\% \\
Test (private) & Expert & 2{,}372 & 5.6 & 85 & 47.2\% \\
\midrule
\textbf{Total} & & \textbf{10{,}173} & \textbf{25.1} & \textbf{857} & \textbf{33.3\%} \\
\midrule
\multicolumn{6}{@{}l}{\textbf{\sbaudio}} \\
\midrule
Train (public) & Expert & 3{,}974 & 10.0 & 312 & 17.2\% \\
Train (public) & Pseudo & 3{,}324 & 9.6 & 447 & 12.8\% \\
Test (private) & Expert & 1{,}094 & 2.4 & 47 & 29.3\% \\
\midrule
\textbf{Total} & & \textbf{8{,}392} & \textbf{22.0} & \textbf{806} & \textbf{17.0\%} \\
\bottomrule
\end{tabular*}
\end{table}

\subsection{Source-disjoint partitions}
\label{sec:splits}
Annotator judgments are aggregated and clip boundaries optimized jointly by dynamic programming into 5--20\,s segments. Compared with sentence-based segmentation (64.5\% label purity, the share of a segment's annotated time that carries its majority label set), this reaches 94.2--98.0\% purity (Supplement~\ref{app:construction}). To prevent speaker and context leakage~\citep{kapoor2023leakage}, the splits are recording-disjoint: the test sets share no recording, channel, or participant with the training data, and are selected at the source level to balance label density, annotation evidence, and engagement (Supplement~\ref{app:partitions}).

\subsection{Annotation}
\label{sec:annotation}
\label{sec:annotation-pipeline}

The expert raters for Social Signal Bench have at least a bachelor's degree in behavioral science or social science and work from a written guide that asks them to annotate a signal only when clear evidence is present. Every expert-annotated clip, including every test clip, receives at least three independent annotations~\citep{sheng2008get}. Crowd votes were added to the trained-rater annotations and count only where a trained rater marked the same or a related signal. This involved only video clips: 7.0\% of the public expert-annotated video set and 6.6\% of the video test set.

Before production work, each rater labels calibration videos that an expert reviews individually. Each label spans at least two seconds, may overlap other labels, and carries an intensity (1--5) and a confidence (Low, Medium, High). The two ratings are independent: a subtle cue can be unambiguous, and a strong display can be ambiguous between signals. When behavior sits between two signals, raters may apply both at low confidence.

Disagreement is measured against the taxonomy rather than by exact label match, using the same credit tiers as EW-F1 (Section~\ref{sec:ewf1}): a vote for \emph{hesitation} where peers labeled \emph{uncertainty} is partially supported, whereas one where peers labeled \emph{agreement} is not. Only cluster-severe votes, those with no support in the same or an adjacent cluster, go to review by an internal annotation expert. On an audit cohort, review leaves 68.7--93.4\% of annotations unchanged and moves cluster-weighted disagreement by at most 0.015 (Supplement Table~\ref{tab:review-effects}). Residual disagreement concerns temporal boundaries more than signal presence (Supplement~\ref{sec:disagreement-measurement}), which motivates clip-level evaluation. Inter-rater agreement is lowest for interest ($\kappa=0.19$), skepticism (0.23), stress (0.25), and uncertainty (0.33).

Reviewed votes are merged into time slices. For a signal that $n$ trained raters assert in a slice, with mean confidence weight $\bar a$ (0.33, 0.66, and 1 for Low, Medium, and High), the slice's evidence score is $s_j=1-(1-\bar a)^{n}$ (Supplement~\ref{sec:ground-truth}), which rises with both confidence and repeated support. A reference signal's evidence weight $p_g$ is the overlap-weighted mean of $s_j$ over the slices that intersect the clip (Eq.~\ref{eq:evidence-weight}). A signal marked with high confidence by every rater approaches $p_g=1$, and a low-confidence label from one rater carries little mass. Figure~\ref{fig:teaser} shows this merge on one training window.

\section{Evidence-Weighted F1}
\label{sec:ewf1}
Conventional F1 treats all incorrect predictions equally, even when social signals are closely related or their presence is ambiguous to human observers. Evidence-Weighted F1 (EW-F1) addresses these limitations by giving partial credit for related signals and weighting reference labels by the strength of their annotation evidence.

\paragraph{Graded credit.}
EW-F1 grades the semantic proximity of a predicted signal $q$ to a reference signal $g$:
\begin{equation}
c(q,g)=
\begin{cases}
1 & q=g,\\
\alpha & q\neq g,\ \mathrm{Cluster}(q)=\mathrm{Cluster}(g),\\
\beta & \{\mathrm{Cluster}(q),\mathrm{Cluster}(g)\}\in\mathrm{Adj},\\
0 & \text{otherwise,}
\end{cases}
\label{eq:match-credit}
\end{equation}
with $\alpha=0.66$ for same-cluster and $\beta=0.33$ for adjacent-cluster matches. Labels outside the taxonomy receive no credit. Graded credit by taxonomy distance follows hierarchical classification measures~\citep{kosmopoulos2015evaluation}.

\paragraph{Evidence weight.}
For a reference signal $g$ in a clip, we define the evidence weight:
\begin{equation}
p_g=\frac{\sum_j s_j\,o_j}{\sum_j o_j}\in(0,1],
\label{eq:evidence-weight}
\end{equation}
where $o_j>0$ is the time in seconds that evidence slice $j$ overlaps the clip and $s_j$ is its evidence score (Section~\ref{sec:annotation}).

\paragraph{Corpus objective.}
Within each clip, predictions are paired injectively with reference signals of positive credit. For an assignment $A$ over the corpus, we define the earned credit as $S(A)=\sum_{(q,g)\in A}p_g\,c(q,g)$, the reference mass as $W=\sum_g p_g$, and the prediction mass as $N'(A)=\sum_{(q,g)\in A}p_g+|\{q\text{ unpaired}\}|$. A matched prediction counts with the weight of its reference signal, and an unmatched prediction counts in full, which penalizes overprediction. Our proposed F1-score is then
\begin{equation}
\mathrm{EW\text{-}F1}=\max_{A}\ \frac{2\,S(A)}{N'(A)+W}.
\label{eq:ewf1}
\end{equation}
With $p_g=1$ for every reference signal and $\alpha=\beta=0$, Eq.~\ref{eq:ewf1} reduces to micro-F1. One-to-one matching prevents multiple predictions from receiving credit for the same reference signal, limiting the benefit of predicting many related labels.

\paragraph{Exact optimization.}
Eq.~\ref{eq:ewf1} is a ratio of linear functions of $A$. Dinkelbach's method~\citep{dinkelbach1967nonlinear} solves it exactly as a short sequence of per-clip maximum-weight bipartite matchings~\citep{kuhn1955hungarian} (Supplement~\ref{app:compute}), so the score is an exact corpus quantity, not an average of per-clip F1 values. With other pairs fixed, adding a prediction for an unmatched reference signal raises the score exactly when $c(q,g)>\mathrm{EW\text{-}F1}/2$, and adding an unrelated prediction never does.

Figure~\ref{fig:matching} illustrates four different model scores for a clip whose reference contains \emph{uncertainty} ($p_g=0.9$) and \emph{hesitation} ($p_g=0.6$). Exact F1 prefers a missing prediction of \emph{uncertainty} to prediction of its sibling \emph{confusion}; EW-F1 reverses that preference. Supplement Table~\ref{tab:ewf1-example} lists $S(A)$ and $N'(A)$ for each answer.

\begin{figure}[t]
\centering
\begin{minipage}[b]{0.48\textwidth}\centering
\resizebox{\linewidth}{!}{\begin{tikzpicture}[
    font=\scriptsize,
    chip/.style={rounded corners=1.5pt, minimum width=1.55cm, minimum height=0.36cm, inner sep=1pt},
    exact/.style={chip, fill=cvprblue, text=white},
    sibling/.style={chip, fill=cvprblue!35},
    missed/.style={chip, draw=black!45, densely dashed, text=black!50},
    unmatched/.style={chip, draw=black!60, fill=white},
    ref/.style={chip, text=white},
    hdr/.style={text=black!60, inner sep=0pt},
    ewbar/.style={fill=cvprblue},
    exbar/.style={fill=ssOrange},
]
  \def\U{0.8}  \def\H{2.45}  \def\X{4.1}   %
  \def\B{5.05} \def\S{2.0}                  %

  \fill[black!6, rounded corners=2pt] (-0.05,-1.33) rectangle (8.0,-2.67);
  \node[align=center, text=black!60, inner sep=0pt] at (7.58,-2.0) {order\\flips};

  \node[hdr] at ({(\U+\H)/2},0.5) {reference};
  \node[hdr] at (\X,0.5) {unmatched};
  \node[ref, fill=black!80] at (\U,0.1) {uncertainty};
  \node[ref, fill=black!50] at (\H,0.1) {hesitation};
  \node at (\U,-0.27) {$p_g{=}0.9$};
  \node at (\H,-0.27) {$p_g{=}0.6$};
  \fill[ewbar] (\B,0.36) rectangle ++(0.3,0.13);
  \node[anchor=west, inner sep=1pt] at (\B+0.3,0.425) {EW-F1};
  \fill[exbar] (\B,0.04) rectangle ++(0.3,0.13);
  \node[anchor=west, inner sep=1pt] at (\B+0.3,0.105) {exact F1};
  \draw[black!30] (-0.05,-0.55) -- (8.0,-0.55);

  \foreach \y/\u/\ustyle/\x/\ew/\ex in {
      -0.97/uncertainty/exact//1.000/1.000,
      -1.65/confusion/sibling//0.796/0.500,
      -2.33/missed/missed//0.571/0.667,
      -3.01/missed/missed/boredom/0.387/0.500} {
    \node[\ustyle] at (\U,\y) {\u};
    \node[exact] at (\H,\y) {hesitation};
    \if\relax\detokenize\expandafter{\x}\relax\else
      \node[unmatched] at (\X,\y) {\x};
    \fi
    \fill[ewbar] (\B,\y+0.04) rectangle (\B+\S*\ew,\y+0.2);
    \node[anchor=west, inner sep=1pt] at (\B+\S*\ew,\y+0.12) {\ew};
    \fill[exbar] (\B,\y-0.2) rectangle (\B+\S*\ex,\y-0.04);
    \node[anchor=west, inner sep=1pt, text=ssOrange] at (\B+\S*\ex,\y-0.12) {\ex};
  }
  \draw[black!40] (\B,-0.68) -- (\B,-3.3);

  \foreach \x/\s/\t in {0.0/exact/{exact: 1}, 1.55/sibling/{same cluster: 0.66}} {
    \node[\s, minimum width=0.4cm, minimum height=0.22cm] at (\x+0.2,-3.65) {};
    \node[anchor=west, inner sep=1pt] at (\x+0.42,-3.65) {\t};
  }
  \node[unmatched, minimum width=0.4cm, minimum height=0.22cm] at (4.25,-3.65) {};
  \node[anchor=west, inner sep=1pt] at (4.47,-3.65) {no credit, adds 1 to $N'$};
\end{tikzpicture}}
\end{minipage}\hfill
\begin{minipage}[b]{0.48\textwidth}\centering
\resizebox{\linewidth}{!}{\begin{tikzpicture}[x=14.2cm, y=7.75cm, font=\scriptsize]
  \def\ylo{0.66}

  \draw[->] (0.08,0) -- (0.53,0);
  \node[below=16pt] at (0.30,0) {perturbation rate $r$};
  \draw[->] (0.08,0) -- (0.08,0.36)
    node[above right=-1pt] {score};

  \foreach \x in {0.1,0.2,0.3,0.4,0.5}
    \draw (\x,0) -- (\x,-0.012) node[below] {\x};

  \foreach \y in {0.7,0.8,0.9,1.0}
    \draw (0.08,\y-\ylo) -- (0.073,\y-\ylo)
      node[left] {\y};

  \draw[thick, cvprblue]
    (0.1,0.979-\ylo) --
    (0.2,0.959-\ylo) --
    (0.3,0.938-\ylo) --
    (0.4,0.922-\ylo) --
    (0.5,0.903-\ylo)
    node[right, xshift=6pt] {same};

  \draw[thick, cvprblue, dash pattern=on 3pt off 1.5pt]
    (0.1,0.961-\ylo) --
    (0.2,0.923-\ylo) --
    (0.3,0.882-\ylo) --
    (0.4,0.850-\ylo) --
    (0.5,0.812-\ylo)
    node[right, xshift=6pt, yshift=4pt] {adj.};

  \draw[thick, cvprblue, dotted]
    (0.1,0.933-\ylo) --
    (0.2,0.864-\ylo) --
    (0.3,0.796-\ylo) --
    (0.4,0.739-\ylo) --
    (0.5,0.673-\ylo)
    node[right, xshift=6pt] {unrel.};

  \draw[thick, black!40]
    (0.1,0.969-\ylo) --
    (0.2,0.937-\ylo) --
    (0.3,0.899-\ylo) --
    (0.4,0.863-\ylo) --
    (0.5,0.826-\ylo)
    node[right, xshift=6pt, yshift=-4pt] {delete};

  \draw[thick, ssOrange]
    (0.1,0.946-\ylo) --
    (0.2,0.893-\ylo) --
    (0.3,0.839-\ylo) --
    (0.4,0.793-\ylo) --
    (0.5,0.743-\ylo);

  \draw[thick, ssOrange, dash pattern=on 3pt off 1.5pt]
    (0.1,0.942-\ylo) --
    (0.2,0.883-\ylo) --
    (0.3,0.823-\ylo) --
    (0.4,0.774-\ylo) --
    (0.5,0.716-\ylo);

  \draw[thick, ssOrange, dotted]
    (0.1,0.942-\ylo) --
    (0.2,0.880-\ylo) --
    (0.3,0.818-\ylo) --
    (0.4,0.766-\ylo) --
    (0.5,0.704-\ylo);

  \node[ssOrange, anchor=west]
    at (0.12,0.695-\ylo)
    {exact F1, three substitutions};

\end{tikzpicture}}
\end{minipage}

\medskip
\begin{minipage}[t]{0.48\textwidth}
\caption{\textbf{Example.} Each prediction sits under its matched reference signal, colored by credit. Exact F1 prefers omitting \emph{uncertainty} (0.667) to predicting its sibling \emph{confusion} (0.500); EW-F1 reverses this (0.571 vs.\ 0.796). Out-of-taxonomy \emph{boredom} earns no credit and adds 1 to $N'(A)$.}
\label{fig:matching}
\end{minipage}\hfill
\begin{minipage}[t]{0.48\textwidth}
\caption{EW-F1 (blue) and exact F1 (orange) on the \sbvideo test reference as labels are replaced by a same-cluster (solid), adjacent (dashed), or unrelated (dotted) signal, or deleted (grey; the two metrics coincide). Each point averages 20 random draws.}
\label{fig:perturbation}
\end{minipage}
\end{figure}

Clips from one recording are dependent, so EW-F1 intervals subsample parent recordings without replacement, not clips~\citep{field2007bootstrapping, politis1994large}; Supplement~\ref{app:errorbars} gives the procedure. Paired comparisons use the same draws for both systems. Signals supported by fewer than 20 parents are reported as not estimable.

\section{Validating EW-F1}
\label{sec:validation}

A new metric needs evidence that it behaves as intended and that its conclusions do not depend on hand-set parameters. All analyses use the full 2{,}372-clip \sbvideo test set: perturbations need only its reference, and parameter tests rescore the eight leaderboard systems of Section~\ref{sec:results} (Supplement~\ref{sec:supp-validation}). Perturbation scores use the 0--1 scale.

We start from a perfect prediction, the reference itself, and damage a fraction $r$ of its labels in one of five ways. A label is replaced by a signal from the same cluster, an adjacent cluster, or an unrelated cluster, or it is deleted, or an out-of-taxonomy label is inserted next to it. Each setting is averaged over 20 random draws (Figure~\ref{fig:perturbation}). EW-F1 orders the three substitutions as the taxonomy intends, at every rate. At $r=0.5$ it scores 0.903 for same-cluster, 0.812 for adjacent, and 0.673 for unrelated substitutions. Exact F1 compresses the same three errors into a band of 0.039 (0.743, 0.716, 0.704) and ranks all of them below deletion (0.825). Exact F1 therefore teaches a system to withhold a near miss. EW-F1 rewards a same-cluster guess over deletion (0.903 versus 0.826), scores an adjacent guess close to deletion (0.812), and penalizes an unrelated guess more than deletion.

The tiers $\alpha=0.66$ and $\beta=0.33$ are hand-set. We rescore the eight-system \sbvideo leaderboard on the grid $\alpha\in\{0.5,0.66,0.8\}\times\beta\in\{0.17,0.33,0.5\}$. The post-trained model stays first and the three lowest systems keep their ranks in all nine cells; only the four frontier models within 0.5 points of each other swap places, and no score moves by more than 3.1 points. Graded credit also does not inflate scores: exact matches supply 89.2\% of the post-trained model's earned credit (Supplement~\ref{sec:supp-leaderboard-metrics}).

Setting $p_g=1$ for every label leaves the post-trained model's score unchanged (44.0), raises the other leaderboard scores by up to 2.1 points, and reorders only the four frontier models within 0.5 points (Kendall's $\tau=0.57$~\citep{kendall1938new}). The weights thus settle near ties in favor of labels that human evidence supports; exact micro- and macro-F1 give $\tau=0.71$ and 0.57 (Supplement~\ref{sec:supp-leaderboard-metrics}).

\section{Reference Results}
\label{sec:results}
\label{sec:experiments}

We evaluate frontier multimodal models and two post-trained open-weight models (Qwen2.5-Omni-7B for \sbvideo, Qwen2-Audio-7B-Instruct for \sbaudio; Supplement~\ref{sec:supp-post-training}), and compare model predictions with individual human judgments.

\paragraph{Setup.}
Within each modality, all systems receive identical test clips, prompts, output contracts, and scoring. A shared harness gives label definitions, observable cues, abstention instructions, and a structured output format; models without native video-with-audio input receive sampled frames and a transcript (Supplement~\ref{app:harness}). Models are trained and evaluated separately on \sbvideo (all ten signals and engagement status) and \sbaudio, each with independently collected annotations. Unlike \sbvideo, \sbaudio focuses on six social signals, excluding \textit{confusion, skepticism, stress}, and \textit{disagreement}. These four signals were insufficiently represented by clear, identifiable instances in the audio-only annotations, making their reliable evaluation difficult. In particular, some of these signals may depend on visual or broader conversational cues that are not consistently recoverable from audio alone. For consistency, annotations of the four excluded signals are removed from the audio-only test references before scoring. Consequently, evaluation is conducted separately within each modality, and scores should not be interpreted as directly comparable across modalities.

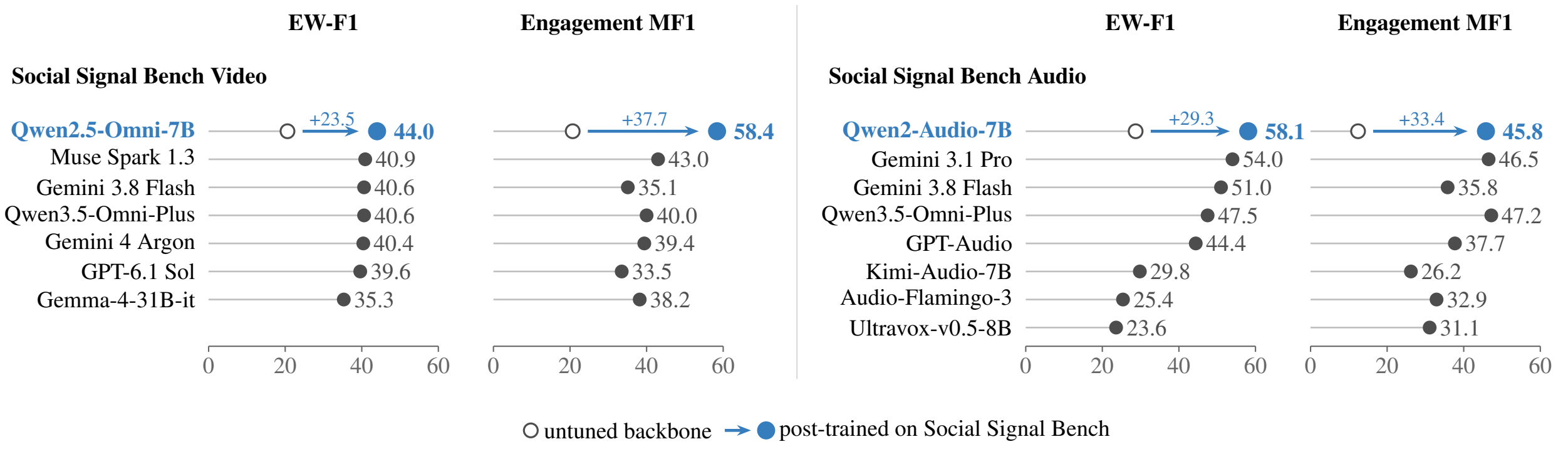
\begin{figure}[t]
\centering
\resizebox{\textwidth}{!}{\begin{tikzpicture}[
    font=\scriptsize,
    sys/.style={circle, fill=black!70, inner sep=0pt, minimum size=4.2pt},
    base/.style={circle, draw=black!70, fill=white, line width=0.6pt, inner sep=0pt, minimum size=4.2pt},
    ours/.style={circle, fill=cvprblue, inner sep=0pt, minimum size=5.5pt},
    sysname/.style={anchor=east, inner sep=1pt},
    val/.style={anchor=west, inner sep=1pt, text=black!70},
    stem/.style={black!25, line width=0.5pt},
    gain/.style={-{Stealth[length=1.5mm]}, cvprblue, line width=0.9pt, shorten >=2.6pt, shorten <=2.2pt},
]
  \def\xo{2.15}   \def\xe{5.2}   \def\xs{0.041}   %
  \def\dy{0.3}                                    %

  \def\metric#1#2#3#4#5#6{%
    \draw[stem] (#1,#2-\dy) -- (#1+\xs*#4,#2-\dy);
    \node[base] (u) at (#1+\xs*#4,#2-\dy) {};
    \node[ours] (o) at (#1+\xs*#5,#2-\dy) {};
    \draw[gain] (u) -- (o);
    \node[val, text=cvprblue, font=\scriptsize\bfseries] at ([xshift=1pt]o.east) {#5};
    \node[text=cvprblue, inner sep=1pt, anchor=south, font=\tiny] at ($(u)!0.5!(o)+(0,0.02)$) {#6};
    \foreach \s [count=\i from 2] in {#3} {
      \draw[stem] (#1,#2-\i*\dy) -- (#1+\xs*\s,#2-\i*\dy);
      \node[sys] at (#1+\xs*\s,#2-\i*\dy) {};
      \node[val] at (#1+\xs*\s+0.07,#2-\i*\dy) {\s};
    }
  }
  \def\names#1#2#3#4{%
    \node[anchor=south west, font=\scriptsize\bfseries, inner sep=1pt] at (0,#1+0.12) {#2};
    \node[sysname, text=cvprblue, font=\scriptsize\bfseries] at (\xo-0.1,#1-\dy) {#3};
    \foreach \n [count=\i from 2] in {#4} \node[sysname] at (\xo-0.1,#1-\i*\dy) {\n};
  }

  \def\ax{-8*\dy-0.2}
  \def\panelframe{%
    \node[anchor=base, font=\scriptsize\bfseries, inner sep=1pt] at (\xo+\xs*30,0.78) {EW-F1};  %
    \node[anchor=base, font=\scriptsize\bfseries, inner sep=1pt] at (\xe+\xs*30,0.78) {Engagement MF1};
    \foreach \xx in {\xo,\xe} {
      \draw[black!60] (\xx,\ax) -- (\xx+\xs*60,\ax);
      \foreach \t in {0,20,40,60}
        \draw[black!60] (\xx+\xs*\t,\ax) -- ++(0,-0.06) node[below, inner sep=1.5pt] {\t};
    }
  }

  \begin{scope}
    \panelframe
    \names{0}{Social Signal Bench Video}{Qwen2.5-Omni-7B}{Muse Spark 1.3, Gemini 3.8 Flash, Qwen3.5-Omni-Plus,
           Gemini 4 Argon, GPT-6.1 Sol, Gemma-4-31B-it}
    \metric{\xo}{0}{40.9, 40.6, 40.6, 40.4, 39.6, 35.3}{20.6}{44.0}{+23.5}
    \metric{\xe}{0}{43.0, 35.1, 40.0, 39.4, 33.5, 38.2}{20.7}{58.4}{+37.7}
  \end{scope}

  \begin{scope}[xshift=8.75cm]
    \panelframe
    \names{0}{Social Signal Bench Audio}{Qwen2-Audio-7B}{Gemini 3.1 Pro, Gemini 3.8 Flash, Qwen3.5-Omni-Plus,
           GPT-Audio, Kimi-Audio-7B, Audio-Flamingo-3, Ultravox-v0.5-8B}
    \metric{\xo}{0}{54.0, 51.0, 47.5, 44.4, 29.8, 25.4, 23.6}{28.7}{58.1}{+29.3}
    \metric{\xe}{0}{46.5, 35.8, 47.2, 37.7, 26.2, 32.9, 31.1}{12.4}{45.8}{+33.4}
  \end{scope}

  \draw[black!15] (8.45,1.05) -- (8.45,\ax-0.35);

  \def\ly{\ax-0.9}
  \begin{scope}[xshift=5.5cm]
    \node[base] at (0.1,\ly) {};
    \node[anchor=west, inner sep=1pt] at (0.2,\ly) {untuned backbone};
    \draw[gain] (2.1,\ly) -- (2.55,\ly);
    \node[ours] at (2.62,\ly) {};
    \node[anchor=west, inner sep=1pt] at (2.72,\ly) {post-trained on Social Signal Bench};
  \end{scope}
\end{tikzpicture}}
\caption{\textbf{Reference results.} EW-F1 and engagement macro-F1 on the private test sets of \sbvideo (left) and \sbaudio (right) under the shared harness and evaluator. The first row of each panel joins the untuned backbone to the same model post-trained on Social Signal Bench. The datasets use different signal subsets, so scores compare only within a panel. Supplement Table~\ref{tab:frontier} lists full model names and access.}
\label{fig:leaderboard}
\end{figure}

On \sbvideo (Figure~\ref{fig:leaderboard}), the strongest closed and open-weight frontier models, Muse Spark 1.3 and Qwen3.5-Omni-Plus, reach 40.9 and 40.6 EW-F1, five of the six frontier models lie between 39.6 and 40.9, and none reaches 45.0 despite detailed task instructions. On \sbaudio, with a different label set, Gemini 3.1 Pro leads with 54.0, followed by Gemini 3.8 Flash at 51.0. The best open-weight audio baseline, Kimi-Audio-7B-Instruct, reaches 29.8.

Task-specific training improves performance on both datasets. On \sbvideo, post-trained Qwen2.5-Omni-7B reaches 44.0 EW-F1 [41.0, 47.0], $+23.5$ [$+20.8$, $+25.8$] over its untuned backbone and $+3.2$ [$+0.8$, $+5.9$] over the strongest frontier model (paired, 85 recordings). Against each of the six frontier models, the paired difference ranges from $+3.2$ to $+8.8$, and every interval excludes zero. It ranks first under EW-F1 and exact micro-F1 and third under exact macro-F1 (Supplement~\ref{sec:supp-leaderboard-metrics}). By setting, it scores 43.7 on online video, 40.8 on sales calls, and 47.5 on dyadic conversations; the last two rest on 6 and 4 recordings (Supplement~\ref{sec:supp-leaderboard-metrics}). On \sbaudio, post-trained Qwen2-Audio-7B-Instruct reaches 58.1 EW-F1 [49.0, 62.1], $+29.3$ [$+22.8$, $+33.3$] over its untuned backbone and $+4.1$ [$-0.1$, $+6.1$] over the strongest frontier model (paired, recording-level); this interval includes zero, so on audio the post-trained model matches the strongest frontier model rather than clearly exceeding it. It ranks first on audio under EW-F1, exact micro- and macro-F1, and uniform evidence weights.

On the engagement head, the post-trained video model reaches 58.4 macro-F1, against at most 43.0 for frontier models and 20.7 for its untuned backbone; the audio model reaches 45.8, against at most 47.2.

On all 2{,}372 video test clips, muting the audio lowers post-trained Qwen2.5-Omni-7B from 44.0 to 22.1 EW-F1 (paired $-22.0$ [$-25.7$, $-16.9$]) and black frames lower it to 15.4 ($-28.7$ [$-31.9$, $-24.2$]); engagement F1 survives muting (63.5) but not black frames (14.1). Both inputs are out of distribution for the model (Supplement~\ref{sec:supp-modality}).

\paragraph{Human raters.} \label{par:human_raters}

We also score individual trained raters against the frozen reference clips. Unlike during annotation, they see only the clip the models receive. On video, five raters reach a mean EW-F1 of 38.9 [34.9, 42.6] on 313 test clips (36 recordings). On audio, four raters who hear only the audio of 391 \sbvideo test clips (40 recordings) reach 40.6 [36.8, 44.5]. Video rater scores fall as clips carry more signals, while audio rater scores are highest on single-signal clips. On clips with no reference signal, raters select nothing in only 47.4\% (video) and 32.2\% (audio) of cases (Supplement~\ref{app:human-reference}). Because the reference pools several raters, an individual rater can score below a frontier model; we report the rater mean as one human perspective, not as a ceiling, and do not rank it against models.

\section{Discussion and Limitations}
\label{sec:discussion}
\label{sec:limitations}
Social Signal Bench labels reflect perceptions of trained expert observers rather than clinical or ground-truth mental states. A high score therefore indicates agreement with human judgments, not necessarily access to a speaker's underlying intentions or emotions. Human baseline performance provides an important reference in that even human observers do not perfectly recover the annotated signals from isolated clips, highlighting the ambiguity of social perception and the role of conversational context. Raters annotated with the surrounding conversation in view, whereas systems and the human baseline see isolated clips, so some reference labels may not be recoverable from the clip alone. Consequently, the gap between model and human performance should be interpreted alongside this difference in available context. EW-F1 further reflects the strength of positive annotation evidence rather than consensus: weights increase with the number and confidence of covering spans but do not penalize raters who omit a signal (Supplement~\ref{app:evidence-weight}). Thus, missing a weakly supported signal carries less cost than missing one supported repeatedly with high confidence.

\paragraph{Limitations.}
Social signals are often subtle, overlapping and dependent on context, making their interpretation inherently subjective. The two modalities also differ in their observable evidence and label coverage, so their scores are not directly comparable. The credit tiers are versioned parameters, and rankings are stable across the tested values (Section~\ref{sec:validation}). Whether taxonomy-defined severities match human judgments of perceptual similarity remains unvalidated (Supplement~\ref{sec:ewf1-validation}). The parameter analyses use only the \sbvideo test set, and model results come from single runs without seed variance. All four domains are English-language conversation with one clearly visible speaker, limiting generalization to multilingual and multi-party interactions. The dyadic and sales domains contain few test recordings (4 and 6 in \sbvideo), resulting in wide per-domain intervals. Within a test set, grouping by recording does not establish independence across participants appearing in multiple sessions.

\section{Ethics and Data Access}
\label{sec:ethics}
\label{sec:availability}

Social Signal Bench contains identifiable faces, voices, and opinions. To respect consent and redistribution restrictions, only CC~BY~3.0 online video is included in the public training sets, with 10 hours of expert annotations per modality.

Annotations were collected through a structured internship program for behavioral- and social-science students (approximately 12 hours per week over six-month terms), including calibration and expert feedback.

Automated social perception may support coaching and interaction research but also poses risks in surveillance, hiring, and workplace monitoring. Release terms prohibit non-consensual surveillance and high-stakes decisions about individuals. Deployment should require informed consent, subgroup evaluation, and human oversight.

\section{Conclusion}
\label{sec:conclusion}
Social Signal Bench measures whether audiovisual and audio-only models perceive co-occurring social signals as trained experts do. It keeps annotator disagreement as graded evidence, enforces recording-disjoint partitions, and scores models with EW-F1, which credits related predictions in proportion to human support. Frontier models reach at most 40.9 (video) and 54.0 (audio) EW-F1, and 7B open models post-trained on the benchmark surpass or match the strongest of them on the benchmark.

\section*{LLM Usage Statement}
The authors designed and ran the study, collected the annotations, and wrote the paper; LLM-based assistants were used to edit, rewrite, and reformat text, \LaTeX{} and figure code, and analysis scripts, and the authors checked all citations against publisher records. The pseudo-labels in the public training sets were produced by teacher models that are fine-tuned multimodal LLMs (Section~\ref{sec:benchmark}). No LLM produced any label in the expert reference or the test sets.

{\small
\bibliographystyle{ieeenat_fullname}
\bibliography{references}
}

\appendix
\setcounter{figure}{0}
\setcounter{table}{0}
\renewcommand{\thefigure}{S\arabic{figure}}
\renewcommand{\thetable}{S\arabic{table}}
\renewcommand{\theHfigure}{S\arabic{figure}}
\renewcommand{\theHtable}{S\arabic{table}}
\phantomsection
\pdfbookmark[0]{Supplementary Material}{supplementary}
\begin{center}
{\LARGE\scshape Supplementary Material}
\end{center}
\vspace{0.5em}
\suppressfloats[t] %
\noindent The sections of this supplementary material follow the order of the main paper: how the reference is built (Sections~\ref{sec:appendix-annotation}--\ref{app:partitions}), the EW-F1 metric and its validation (Sections~\ref{app:ewf1-derivation}--\ref{sec:supp-validation}), and the evaluation (Sections~\ref{app:harness}--\ref{app:human-reference}).

\section{Labeling}
\label{sec:appendix-annotation}
\label{sec:annotation-methodology}

\begin{figure*}[t]
\centering
\begin{tikzpicture}[
    font=\scriptsize,
    panel/.style={fill=black!3, rounded corners=4pt},
    stepn/.style={circle, fill=cvprblue, text=white, font=\scriptsize\bfseries, inner sep=0pt, minimum size=11pt},
    ttl/.style={anchor=west, font=\small\bfseries, inner sep=1pt},
    note/.style={anchor=north west, text=black!60, align=left, inner sep=1pt, text width=3.6cm},
    span/.style={rounded corners=1.5pt, minimum height=0.27cm, inner sep=0pt, anchor=west},
    lane/.style={black!20, line width=0.4pt},
    flow/.style={-{Stealth[length=2.2mm]}, black!45, line width=1pt},
    flag/.style={draw=ssOrange, dashed, line width=0.8pt},
]
  \colorlet{unc}{cvprblue}
  \colorlet{agr}{ssOrange}
  \colorlet{ssev}{cvprblue}
  \def\W{3.95}  \def\H{3.55}  \def\G{0.42}
  \foreach \i in {0,...,3} \fill[panel] (\i*\W+\i*\G,0) rectangle ++(\W,\H);
  \foreach \i in {0,...,2} \draw[flow] (\i*\W+\i*\G+\W+0.04,\H/2) -- ++(\G-0.08,0);

  \begin{scope}[shift={(0,0)}]
    \node[stepn] at (0.3,\H-0.3) {1};
    \node[ttl] at (0.5,\H-0.3) {Raters label spans};
    \fill[black!8] (0.55+3*0.52,1.15) rectangle (0.55+4*0.52,2.75);
    \node[text=black!50, font=\tiny] at (0.55+3.5*0.52,1.03) {bin $t$};
    \foreach \y/\r in {2.5/R1,2.0/R2,1.5/R3} {
      \draw[lane] (0.55,\y) -- (3.75,\y);
      \node[anchor=east, text=black!60] at (0.52,\y) {\r};
    }
    \node[span, fill=unc, text=white, minimum width=1.82cm] at (0.55+0.5*0.52,2.5) {\,hesitation (H)};
    \node[span, fill=unc!66, minimum width=1.82cm] at (0.55+1.0*0.52,2.0) {\,uncertainty (M)};
    \node[span, fill=agr!40, minimum width=1.82cm] at (0.55+2.0*0.52,1.5) {\,agreement (L)};
    \node[note] at (0.2,0.85) {Spans of at least 2\,s; each carries a confidence L/M/H\,=\,0.33/0.66/1.};
  \end{scope}

  \begin{scope}[shift={(\W+\G,0)}]
    \node[stepn] at (0.3,\H-0.3) {2};
    \node[ttl] at (0.5,\H-0.3) {Measure peer support};
    \node[span, fill=unc, text=white, minimum width=1.25cm, anchor=center] (h) at (0.95,2.5) {hesitation};
    \node[span, fill=unc!66, minimum width=1.25cm, anchor=center] (u) at (0.95,2.0) {uncertainty};
    \node[span, fill=agr!40, minimum width=1.25cm, anchor=center] (a) at (0.95,1.5) {agreement};
    \draw[cvprblue, line width=0.7pt] (h.east) to[bend left=50] node[right, font=\tiny, inner sep=1pt] {0.66} (u.east);
    \draw[black!40, line width=0.5pt, densely dotted] (h.west) to[bend right=50] node[left, font=\tiny, inner sep=1pt, xshift=-2pt] {0} (a.west);
    \node[anchor=west, align=left, inner sep=1pt] at (2.05,2.35) {$S_{\mathrm{hes}}=0.33$\\[-1pt]{\tiny\color{black!60}$(0.66+0)/2$}};
    \node[anchor=west, align=left, inner sep=1pt, text=ssOrange] at (2.05,1.55) {$S_{\mathrm{agr}}=0$\\[-1pt]{\tiny cluster-severe}};
    \node[note] at (0.2,0.85) {One-second bins. Support is the mean best taxonomy credit (1/0.66/0.33/0) from each peer.};
  \end{scope}

  \begin{scope}[shift={(2*\W+2*\G,0)}]
    \node[stepn] at (0.3,\H-0.3) {3};
    \node[ttl] at (0.5,\H-0.3) {Expert reviews flags};
    \node[span, fill=agr!40, minimum width=1.25cm, anchor=center, flag] (f) at (0.95,2.05) {agreement};
    \node[draw=black!50, fill=white, rounded corners=2pt, inner sep=2.5pt, anchor=west] (e) at (2.05,2.05) {expert};
    \draw[flow, black!55, line width=0.7pt] (f) -- (e);
    \node[anchor=north, align=center, text=black!70, inner sep=1pt, font=\tiny] at (e.south) {keep, remove, relabel,\\re-time, re-rate};
    \node[note] at (0.2,0.85) {Only cluster-severe votes are \mbox{reviewed}. Audit: 68.7--93.4\% of annotations unchanged.};
  \end{scope}

  \begin{scope}[shift={(3*\W+3*\G,0)}]
    \node[stepn] at (0.3,\H-0.3) {4};
    \node[ttl] at (0.5,\H-0.3) {Merge into evidence};
    \node[anchor=west, inner sep=1pt] at (0.25,2.82) {$s_j=1-(1-\bar a)^{n}$};
    \def\bx{1.55} \def\bs{1.85}
    \foreach \lab/\v [count=\k] in {1$\times$Low/0.33, 1$\times$Medium/0.66, 3$\times$Low/0.70, High$+$Medium/0.97, 3$\times$High/1.00} {
      \pgfmathsetmacro{\yy}{2.5-0.27*\k+0.27}
      \node[anchor=east, inner sep=1pt, font=\tiny] at (\bx-0.05,\yy) {\lab};
      \fill[ssev] (\bx,\yy-0.09) rectangle ++(\bs*\v,0.18);
      \node[anchor=west, inner sep=1pt, font=\tiny, text=black!70] at (\bx+\bs*\v,\yy) {\v};
    }
    \node[note] at (0.2,0.85) {$n$ raters with mean confidence $\bar a$ per time slice; $p_g$ averages $s_j$ over the clip.};
  \end{scope}
\end{tikzpicture}
\caption{\textbf{From rater spans to the evidence-weighted reference} (schematic). (1)~Trained raters label spans independently, each with a confidence. (2)~In every one-second bin, each vote earns support from peers under the taxonomy credit; a vote with no support in its own or an adjacent cluster is cluster-severe. (3)~An expert reviews only cluster-severe votes. (4)~Reviewed votes are merged per signal and time slice; the slice score rises with both confidence and the number of raters, and a clip's evidence weight $p_g$ is its overlap-weighted mean.}
\label{fig:supp-labeling}
\end{figure*}
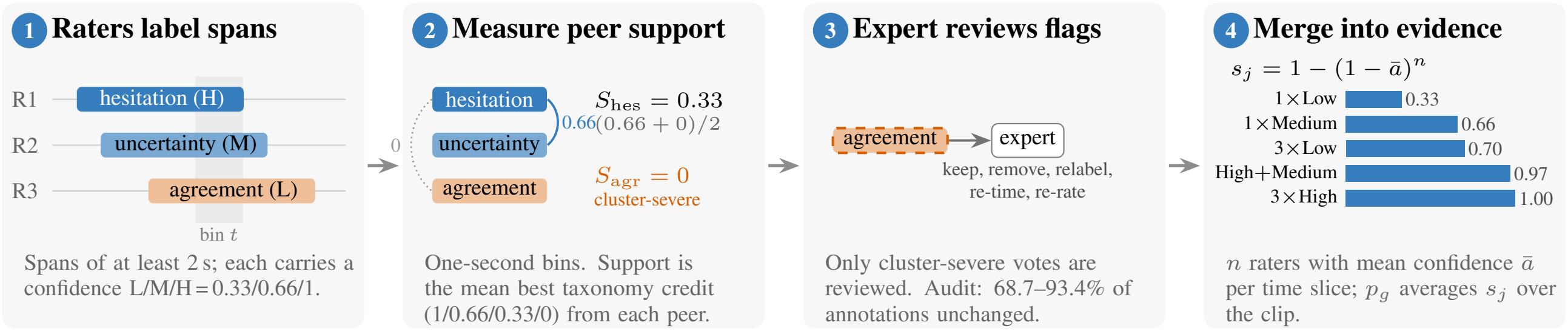

Figure~\ref{fig:supp-labeling} summarizes how independent rater spans become the reference.

\subsection{Annotation protocol}
Annotators label a signal only when it is clearly observable, and they are told that a false positive is worse than a gap. Annotation is restricted to English speech and to a single person who is clearly visible and speaking; static background people who do not interfere are the only exception. Each span lasts at least two seconds, and each judgment uses both the visual and the audio channel. Several labels may be active over the same span, because social signals often co-occur and a single label per moment would misrepresent them.

Each label carries two ratings. Intensity (1--5) records how strongly the signal is displayed, and confidence (Low, Medium, High) records how certain the annotator is that the label is correct. The two ratings are independent: a high-arousal behavior can carry low confidence when its cause is ambiguous between signals, and a subtle cue can carry high confidence when it is unambiguous. When a behavior sits between two signals, annotators may apply both at low confidence. Low confidence still requires an observed behavior that is attributable to at least one of the twelve signals; if nothing in the taxonomy applies, no label is applied. Intensity is collected but not yet used in the disagreement, review, or ground-truth computations below (Section~\ref{sec:annotation-limitations}).

\subsection{Disagreement and review}
\label{sec:disagreement-measurement}
All disagreement measures share one temporal discretization. Spans are binned into one-second intervals, and a label is active in every bin that its span covers. When two overlapping spans of the same label from the same annotator fall in one bin, the higher confidence is kept. Confidence maps to the weights Low~=~0.33, Medium~=~0.66, and High~=~1.00 throughout.

Signals are grouped into seven clusters (Agreement; Confidence; Disengagement; Engagement \& Interest; Skepticism \& Disagreement; Stress \& Frustration; Uncertainty, Hesitation \& Confusion). The clusters form a fixed adjacency graph: \{Agreement, Confidence, Engagement \& Interest\} form one triangle, \{Skepticism \& Disagreement, Stress \& Frustration, Uncertainty \& Hesitation\} form another, and Disengagement is adjacent to no cluster. This gives a similarity weight between any two labels:

\begin{equation}
w(\ell, k) =
\begin{cases}
1.00 & k = \ell \\
0.66 & k \neq \ell,\ \text{same cluster} \\
0.33 & \text{adjacent clusters} \\
0.00 & \text{otherwise}
\end{cases}
\end{equation}

These weights encode the intended structure of the taxonomy; they are set by hand, not estimated from data (Section~\ref{sec:annotation-limitations}). Table~\ref{tab:signal-taxonomy} lists each signal with its cluster and the perceived behavior that defines it.

\begin{table}[h]
\centering
\caption{The twelve-signal taxonomy, its seven clusters, and the perceived behavior that defines each signal. Engagement and Disengagement form a separate engagement head; EW-F1 scores the other ten signals.}
\label{tab:signal-taxonomy}
\tablefont
\begin{tabularx}{\textwidth}{@{}lp{0.26\textwidth}>{\raggedright\arraybackslash}X@{}}
\toprule
\textbf{Signal} & \textbf{Cluster} & \textbf{Perceived behavior} \\
\midrule
Agreement & Agreement & Visible alignment or affirmation \\
Confidence & Confidence & Assured, steady stance toward one's message \\
Disengagement & Disengagement & Withdrawal of attention or involvement \\
Engagement & Engagement \& Interest & Sustained, active connection to task or partner \\
Interest & Engagement \& Interest & Motivated attention to something new \\
Disagreement & Skept.\ \& Disagr. & Open opposition to another's position \\
Skepticism & Skept.\ \& Disagr. & Doubting stance toward another's statement \\
Frustration & Stress \& Frustr. & Annoyance from obstacles or blocked goals \\
Stress & Stress \& Frustr. & Tension and anxiety, effort to stay in control \\
Confusion & Uncert.\ \& Hesit. & Momentary breakdown in understanding \\
Hesitation & Uncert.\ \& Hesit. & Disruption in speaking or responding \\
Uncertainty & Uncert.\ \& Hesit. & Self-doubt, partial commitment to a message \\
\bottomrule
\end{tabularx}
\end{table}

We use two families of disagreement measures. The pairwise family compares the active label sets of each pair of annotators in each bin, by Jaccard overlap (exact matching) or by a soft best-match Jaccard under $w$ (cluster-aware matching), and averages over bins and pairs. The support family scores each annotator's vote by the support it receives from peers: the mean, over peers, of the similarity weight of their best-matching label. Both families have confidence-naive and confidence-weighted variants. The confidence-weighted support variant takes the square root of the support before converting it to a disagreement score, so the first corroborating peer reduces disagreement more than later ones:

\begin{gather}
D_{\text{cluster}}(i, \ell, t) = \left(1 - \sqrt{S_{\text{cluster}}(i, \ell, t)}\right) c(i, \ell, t),\\
S_{\text{cluster}}(i, \ell, t) = \frac{1}{|P_i|}\sum_{j \in P_i} \max_{k \in L_{j,t}} w(\ell, k)
\end{gather}

$D_{\text{exact}}$ is defined in the same way with exact-match support. By default we report the exact pairwise measure (``Disagreement Score'') and the cluster-aware, confidence-weighted support measure (``Cluster-weighted Disagreement Score''). All scores lie in $[0,1]$ and are reported in fixed bands: Low ($0<x\le0.35$), Medium ($\le0.51$), High ($\le0.70$), and Severe ($>0.70$). The thresholds are the same for every variant and are not calibrated to the observed distribution. We also report pairwise Cohen's $\kappa$ per signal, computed on the same bins by treating membership in an annotator's label set as a binary indicator and pooling a 2$\times$2 confusion matrix over all shared bins. Unlike the taxonomy-aware scores, $\kappa$ is a standard chance-corrected reliability statistic.

\label{sec:review-adjudication}
Review is triggered by cluster-severe votes, those with no support in the same or an adjacent cluster (main paper Section~\ref{sec:annotation}). An expert reviewer inspects the flagged clips and may remove a span, add a missing label, relabel a span, or adjust its confidence or temporal boundaries. The aim is to correct clear errors and add missing evidence, not to drive disagreement toward zero. Social signals are fuzzy and depend on context, and a fixed cue catalogue captures them only in part. Disagreement that survives review is therefore not read as a process failure, and a reviewed label is not treated as the single correct reading.

\label{sec:review-audit}
Table~\ref{tab:review-effects} audits review on recordings with both pre- and post-review exports: 865 audiovisual clips and 887 audio-only recordings from online video and dyadic conversations. Review is conservative. It leaves 88.0--93.4\% of audiovisual annotations and 90.3\% of online-video audio annotations unchanged, and it moves cluster-weighted disagreement by at most 0.015. Dyadic audio is revised more (68.7\% unchanged), yet its cluster-weighted disagreement moves only from 0.433 to 0.418. Review thus revises annotations without enforcing consensus. Residual audiovisual disagreement concerns temporal boundaries more than signal presence, which motivates clip-level evaluation. Agreement is lowest for interest ($\kappa=0.19$), skepticism (0.23), stress (0.25), and uncertainty (0.33).

\begin{table}[h]
\centering
\caption{Disagreement before and after expert review on the annotation audit
cohort. Lower disagreement is better.}
\label{tab:review-effects}
\tablefont
\setlength{\tabcolsep}{4pt}
\begin{tabular*}{\textwidth}{@{\extracolsep{\fill}}lrrrr@{}}
\toprule
& \multicolumn{2}{c}{\textbf{Audiovisual}} & \multicolumn{2}{c}{\textbf{Audio-only}} \\
\cmidrule(lr){2-3}\cmidrule(l){4-5}
\textbf{Metric} & \textbf{Online} & \textbf{Dyadic} & \textbf{Online} & \textbf{Dyadic} \\
\midrule
Raw exact-label disagreement & 0.354 & 0.402 & 0.482 & 0.549 \\
Raw cluster-weighted disagr. & 0.272 & 0.342 & 0.376 & 0.433 \\
Reviewed cluster-weighted disagr. & 0.275 & 0.341 & 0.371 & 0.418 \\
Annotations unchanged by review & 88.0\% & 93.4\% & 90.3\% & 68.7\% \\
Cluster-severe votes removed & 19.6\% & 16.2\% & 7.0\% & 18.4\% \\
\bottomrule
\end{tabular*}
\end{table}

\subsection{From votes to the reference}
\label{sec:ground-truth}
The reference comes from a heuristic evidence-accumulation rule, motivated by work on annotator reliability~\citep{raykar2010learning} and by DAiSEE's separately psychologist-labeled reference subset~\citep{gupta2016daisee}. For a signal $S$ in a temporal segment, let $n_S$ be the number of trained annotators who assert it and $c_{j,S} \in \{0.33,0.66,1.00\}$ their confidence weights. Then
\begin{equation}
\bar c_S = \frac{1}{n_S}\sum_{j=1}^{n_S} c_{j,S}, \qquad
C_S = 1-(1-\bar c_S)^{n_S}.
\label{eq:evidence-confidence}
\end{equation}
The score rises with both mean confidence and repeated support. It is a design heuristic, not a fitted Bayesian annotator model.

Crowd raters took part only in the first annotation cycle, which produced the Online video A domain. Their labels enter through a second branch. Each crowd label is weighted by its cluster distance to the trained-annotator labels in the same slice (1 for the same signal, 0.66 for the same cluster, 0.33 for an adjacent cluster), and a crowd label with no related trained-annotator label is discarded. Where both sources are present, the slice score is the mean of the trained and crowd scores (Section~\ref{app:ewf1-derivation}). Later cycles had a stable pool of trained annotators and dropped crowd raters, so the sales-call, dyadic-conversation, and Online video B domains use only the trained-annotator merge of Eq.~\eqref{eq:evidence-confidence}.

\label{sec:quality-control}
Two automated reports check every finalized ground-truth file before use. The first gives the annotated minutes and clip share of each signal, overall and by confidence level, the engagement partition, and the pairwise co-occurrence of signals (overlap and Jaccard index of their merged spans). The second gives the mean span length of each signal by confidence level. It flags, for example, a signal whose low-confidence spans run systematically longer than its high-confidence spans, which can point to a gap in the labeling convention rather than a property of the signal.

\label{sec:downstream-evaluation}
The cluster adjacency and similarity weights of Section~\ref{sec:disagreement-measurement} reappear in EW-F1 (main-paper Section~\ref{sec:ewf1}), but with a different role: there they score model predictions against the merged reference, not annotators against each other.

\subsection{Limitations of the labels}
\label{sec:annotation-limitations}
Annotated videos run from roughly thirty seconds to three minutes, and annotators view the whole video, whereas a model sees only its clip. Labels are therefore made with more context than the model receives. The number of annotators per clip differs between batches (three in some, four in others), and the disagreement and reliability measures do not correct for this, so per-vote support is not comparable in expectation across clips with different annotator counts. The similarity weights (1.0/0.66/0.33/0.0) used in disagreement scoring and in the crowd branch for Online video A are set by hand and have not been validated against independent reviewer judgment; they do not enter the trained-annotator formula. In early cycles, one person could annotate both the audio and the video task of a batch. The merge treats each export as an independent annotator, so such pairs can inflate repeated support in Eq.~\eqref{eq:evidence-confidence}; later cycles restrict each annotator to one modality per batch. Review covers only cluster-severe votes, so most annotations receive no second pass and disagreement below that threshold is not audited. A single primary reviewer adjudicates, and reviewer-specific effects cannot be separated from disagreement between annotators. Intensity ratings are collected but not used in any disagreement, review, or ground-truth computation. Finally, annotators are internship-program participants who work twelve hours a week over a six-month term. No formal IRB or equivalent ethics review was sought, because the work was treated as an internal work task rather than a human-subjects research protocol; sociodemographic data are collected anonymously.

\section{Clip Construction and Diagnostics}
\label{app:construction}
Clips are cut from annotated recordings by joint segmentation. For a candidate interval $I$ of duration $d_I$, let $c_{I,k}$ be the annotated duration of signal $k$ inside $I$ and $m_{I,k}$ the number of merged annotation-run midpoints inside $I$. Keeping and dropping label $k$ cost
\begin{equation}
C^{\mathrm{on}}_{I,k}=d_I-c_{I,k},\qquad C^{\mathrm{off}}_{I,k}=15\,m_{I,k}+c_{I,k}.
\end{equation}
A label is kept when its on-cost is strictly smaller. Dynamic programming chooses a complete, non-overlapping partition into 5--20\,s clips that minimizes the summed cost, with clip count and duration as tie-breakers. Engagement is derived separately, and clips without other labels stay in the dataset. \sbaudio uses the same cost model and constants (5--20\,s clips, label weight 15, missed-signal weight 1). Each dyadic session has a warm-up, three topical discussion phases driven by clarifying, assumption-testing, and challenging prompts, and a wrap-up.

We measure fidelity in three ways. Purity is the fraction of assigned label time supported by a temporal annotation. Signal-time recall is the fraction of annotated time covered by clips that keep the label. Event recall is the fraction of annotated intervals that overlap at least one such clip. These numbers measure how well clips preserve the annotations, not whether the annotations are correct. Table~\ref{tab:construction} reports them for the full annotated corpus of each modality. The segmentation precedes the selection of the public training sets and private test sets, so its clip counts differ from main-paper Table~\ref{tab:release}.

\begin{table}[h]
\centering
\caption{Agreement between clip labels (social signals) and temporal annotations (\%) on the full annotated corpus, before benchmark selection.}
\label{tab:construction}
\tablefont
\setlength{\tabcolsep}{2pt}
\begin{tabular*}{\textwidth}{@{\extracolsep{\fill}}lrccc@{}}
\toprule
\textbf{Construction} & \textbf{Clips} & \textbf{Purity} & \textbf{Time rec.} & \textbf{Event rec.} \\
\midrule
\multicolumn{5}{@{}l}{\textbf{\sbvideo}} \\
\midrule
Sentence timestamps & 4{,}076 & 64.5 & --- & --- \\
\midrule
Joint, online video & 4{,}480 & 94.2 & 98.2 & 96.4 \\
Joint, added domains & 4{,}542 & 98.0 & 99.3 & 97.4 \\
\midrule
\multicolumn{5}{@{}l}{\textbf{\sbaudio}} \\
\midrule
Joint, online video & 6{,}108 & 96.8 & 98.9 & 97.0 \\
Joint, added domains & 3{,}857 & 96.4 & 98.9 & 96.8 \\
\bottomrule
\end{tabular*}
\end{table}

We keep the natural prevalence of signals in every evaluation set. In an internal diagnostic, a model trained on Social Signal Bench reached macro-F1 0.62 on a set balanced by removing negative examples, but 0.44 at natural prevalence, while its AUC barely changed. Balanced sets would therefore overstate performance.

\section{Test-Set Selection}
\label{app:partitions}
The private test sets are chosen at the level of source recordings. Only reference annotations enter the selection; model predictions and scores are not inputs, and each assignment was frozen before any model was trained on the remaining data.

The unit of assignment is a source group, and every clip of a group goes to the same split. Group identity comes from three independent sources of evidence: the platform video ID, the SHA-256 hash of the parent media, and the parent file name after removal of chunk, window, and clip suffixes. Clips whose parent hashes agree must share a group, even when their chunk URIs differ. Contradictory evidence, such as two video IDs for one file, stops the build. A dyadic conversation is one group, not one group per derived segment.

A mixed-integer program over source groups selects the test sets. Its hard constraints require at least one group from each of the four domains, at least 20 groups with each signal, at least 15 groups with disengaged clips, and no group with more than 350 clips. The objective is a weighted sum of absolute deviations from proportional targets for clips, groups, clips per domain, positive groups per signal, signal instances, empty clips, label density, evidence mass, engagement states, and confidence tiers, each divided by its target and by the number of terms of its kind. The CBC solver runs single-threaded with a relative gap of 0.002 under several fixed seeds. We keep the lowest-objective candidate that passes an independent re-audit of every hard constraint, store the assignment as a hashed manifest, and replay later views from it without solving again.

Table~\ref{tab:partitions} summarizes both test sets. Every video signal has at least 21 positive recordings. In the audio test set, Confidence, Hesitation, and Uncertainty have 24--42 positive recordings, and Agreement, Interest, and Frustration have 17--19. Disengaged clips come from 16 video and 3 audio recordings.

\begin{table}[h]
\centering
\caption{Private test sets. \emph{Rec.}: source recordings, with the recordings per domain (online video A and B, sales calls, dyadic conversations). \emph{Empty}: clips with no annotated signal at any probability among the scored signals.}
\label{tab:partitions}
\tablefont
\setlength{\tabcolsep}{2.5pt}
\begin{tabular*}{\textwidth}{@{\extracolsep{\fill}}lrrccccr@{}}
\toprule
\textbf{Test set} & \textbf{Clips} & \textbf{Rec.} & \textbf{A} & \textbf{B} & \textbf{Sales} & \textbf{Dyad.} & \textbf{Empty}\\
\midrule
\sbvideo & 2{,}372 & 85 & 45 & 30 & 6 & 4 & 47.2\% \\
\sbaudio & 1{,}094 & 47 & 24 & 17 & 5 & 1 & 29.3\% \\
\bottomrule
\end{tabular*}
\end{table}

The test sets stay private and are used only through hosted evaluation. Every submission is scored with the same prediction schema, the same versioned EW-F1 scorer, and the same references. The public leaderboard reports aggregate scores without exposing individual test annotations. Submissions are rate-limited and results are versioned to limit overfitting to the test sets. Each run is recorded in a manifest with the split hash, taxonomy version, model identifier, prompt and parser versions, decoding parameters, and raw predictions.

\section{EW-F1: Full Derivation}
\label{app:ewf1-derivation}

This section derives EW-F1 step by step, proves its properties, gives the exact algorithm, and states its limits. The proofs hold for any reference.

\subsection{Notation}
A \emph{clip} is one scored unit, and several clips can come from one parent recording, which is usually 60 seconds long. For one clip, $G$ is the set of reference signals and $Q$ the set of predicted signals. Both hold distinct names. The letters $n$ and $m$ count parent recordings, never signals. \emph{Expert} spans come from trained raters, and \emph{crowd} spans from the crowd raters of the first annotation cycle (Section~\ref{sec:ground-truth}).

\subsection{The two inputs}

The first input is the match credit $c(q,g)\in\{0,\beta,\alpha,1\}$, which grades how close a prediction $q$ is to a reference signal $g$ (main-paper Eq.~\ref{eq:match-credit}). The credit is symmetric, and the released values are $\alpha=0.66$ and $\beta=0.33$.

The second input is the evidence weight. The merge step cuts each recording into time \emph{slices} at every span boundary. For one signal in one slice, let $\bar a_E$ be the mean confidence of the $n_E$ expert spans that cover it, with Low, Medium, and High mapped to 0.33, 0.66, and 1. The expert score is
\begin{equation}
s^E = 1-(1-\bar a_E)^{n_E}.
\label{eq:slice-score}
\end{equation}
Crowd spans use the same form with a reliability factor: $s^C = r\,\bigl(1-(1-\bar a_C)^{n_C}\bigr)$. Here $r=1$ if an expert marked the same signal in the slice, $r=0.66$ if an expert marked a signal of the same cluster, $r=0.33$ if an expert marked a signal of an adjacent cluster, and $r=0$ otherwise, in which case the crowd span is discarded. The slice score $s_j$ is the mean of $s^E$ and $s^C$ when both exist, and the one that exists otherwise. The clip weight $p_g$ is the duration-weighted mean of $s_j$ over the slices of $g$ that overlap the clip (main-paper Eq.~\ref{eq:evidence-weight}). A signal with no overlapping slice is not in $G$.

\label{app:evidence-weight}
The weight measures volume, not agreement. Both scores rise with the number of covering spans, and neither has a term for a rater who did not mark the signal.

\subsection{Constructing the score}
EW-F1 keeps the logic of F1: reward correct detections, charge missed references, and charge extra predictions. It changes two details: a related signal earns partial credit, and a weakly supported reference signal carries less weight.

The score is built in five steps. First, predictions are paired one-to-one with reference signals within each clip. A pair is allowed only if $c(q,g)>0$, pairs never cross clips, and a pair earns $p_g\,c(q,g)$. Second, an assignment $A$ over the corpus earns the credit $S(A)=\sum_{(q,g)\in A}p_g\,c(q,g)$.

Third, we define two masses. The reference mass $W=\sum_g p_g$ counts every reference signal, found or not, so missing a strong label costs more than missing a weak one. The prediction mass is
\begin{equation}
N'(A)=\sum_{(q,g)\in A}p_g + \bigl|\{q \text{ unpaired in } A\}\bigr|.
\end{equation}
A paired prediction carries the weight of its partner, and an unpaired prediction costs a full unit. The asymmetry is deliberate: weak evidence makes a miss cheaper, but it never makes a false positive cheaper.

Fourth, the two masses give an F1 ratio, $F(A)=2S(A)/(N'(A)+W)$. For an exact predictor $S=N'=W$, so $F=1$. Fifth, several pairings can be valid when related signals earn partial credit, and a scorer that chooses a pairing by one rule but reports another quantity can be gamed (Section~\ref{app:alternatives}). EW-F1 therefore reports the largest value of the ratio itself:
\begin{equation}
\mathrm{EW\text{-}F1}=\max_{A\in\mathcal{A}}\frac{2S(A)}{N'(A)+W},
\end{equation}
where $\mathcal{A}$ is the set of allowed clip-local assignments. If $W=0$, the score is 1 with no predictions and 0 otherwise.

\subsection{Properties}

\begin{proposition}[Range]
\label{prop:range}
$0\le\mathrm{EW\text{-}F1}\le1$, and a system that predicts exactly the reference set of every clip scores 1.
\end{proposition}
\begin{proof}
Since $c\le1$ and each $g$ is paired at most once, $S(A)\le\sum_{(q,g)\in A}p_g\le\min\{W,N'(A)\}$. Hence $2S\le N'+W$. For the exact system, pair each prediction with the identical reference signal. Then $c=1$ for every pair and $S=N'=W$.
\end{proof}

\begin{proposition}[No gain from removing a relation]
\label{prop:mono}
Fix $Q$ and $G$. If $E'\subseteq E$ are two sets of allowed pairs, then $\mathrm{EW\text{-}F1}(E')\le\mathrm{EW\text{-}F1}(E)$.
\end{proposition}
\begin{proof}
$F(A)$ depends only on $A$. Every assignment feasible under $E'$ is feasible under $E$, and a maximum over a subset cannot exceed the maximum over the whole set.
\end{proof}

\begin{proposition}[Tie-breaks do not matter]
\label{prop:tie}
All maximizing assignments report the same value.
\end{proposition}
\begin{proof}
The reported number is the maximum value, not a property of one maximizer.
\end{proof}

\begin{proposition}[Cost of a wrong extra label]
\label{prop:fp}
An added prediction that pairs with nothing adds 1 to $N'$ and 0 to $S$, so it strictly lowers the score whenever $S>0$.
\end{proposition}

\begin{proposition}[When a label is worth predicting]
\label{prop:thresh}
Let $F$ be the present score. Add one prediction that pairs with a free reference signal at credit $c$ and weight $p$, with the other pairs fixed. The score rises if and only if $c>F/2$.
\end{proposition}
\begin{proof}
The new value is $2(S+cp)/(N'+p+W)$. It exceeds $2S/(N'+W)$ if and only if $cp(N'+W)>Sp$, that is, $c>S/(N'+W)=F/2$.
\end{proof}

The condition does not contain $p$, so a system has no reason to withhold a weakly supported label. Proposition~\ref{prop:mono} holds for a fixed prediction set. Adding a prediction is a different operation, because it also adds denominator mass, and it can lower the score.

\subsection{Exact computation}
\label{app:compute}
The maximizer of a ratio is not the maximizer of its numerator, so a greedy or numerator-only search is wrong, and full enumeration is too slow. Let $\lambda\ge0$ be a trial ratio and $D(A)=N'(A)+W$. Then
\begin{equation}
\begin{aligned}
S(A)-\lambda D(A)
&= \sum_{(q,g)\in A}
   \bigl[p_g\,c(q,g)-\lambda(p_g-1)\bigr]\\
&\quad -\lambda|Q|-\lambda W.
\end{aligned}
\label{eq:sep}
\end{equation}
The last two terms do not depend on $A$, and the first is a sum over pairs. For fixed $\lambda$, Eq.~\ref{eq:sep} therefore separates into one maximum-weight assignment per clip, with edge value $v(q,g)=p_g\,c(q,g)-\lambda(p_g-1)$. Dinkelbach's method~\citep{dinkelbach1967nonlinear} updates $\lambda$ to the ratio of the current maximizer (Algorithm~\ref{alg:dinkelbach}).

\begin{algorithm}[h]
\caption{Exact evaluation of EW-F1.}
\label{alg:dinkelbach}
\small
\begin{algorithmic}[1]
\Require clips with weights $p$ and credits $c$; tolerance $\tau=10^{-10}$; cap $K=50$
\State $\lambda\gets0$ \Comment{valid lower bound, since $S\ge0$}
\For{$k\gets1$ \textbf{to} $K$}
  \ForAll{clips}
    \State $A_{\mathrm{clip}}\gets\arg\max_A\sum_{(q,g)\in A}v(q,g)$
  \EndFor
  \State $A^\ast\gets\bigcup A_{\mathrm{clip}}$;\; $r\gets S(A^\ast)-\lambda D(A^\ast)$
  \If{$r\le\tau\,D(A^\ast)$}
    \State \Return $2S(A^\ast)/D(A^\ast)$ \Comment{certified maximum}
  \EndIf
  \State $\lambda\gets S(A^\ast)/D(A^\ast)$
\EndFor
\State \textbf{raise} non-convergence
\end{algorithmic}
\end{algorithm}

The function $h(\lambda)=\max_A\{S(A)-\lambda D(A)\}$ is convex and strictly decreasing, because $D(A)>0$. Its single root is the maximum ratio, so the residual test certifies the returned value without a convergence assumption. The tolerance is relative, so it does not depend on corpus size. Each clip holds only a few signals, so each clip assignment is small and is solved exactly.

Per-signal scores reuse the corpus assignment, which is solved once. For a signal $\sigma$, we keep the pairs whose \emph{reference} signal is $\sigma$, add the unpaired predictions named $\sigma$, and recompute the three masses on that subset. Near-miss credit counts toward the reference signal, because that side owns the recall mass. Inside each resampling draw this gives one score per signal at no extra cost.

\subsection{Why the score must be a maximum}
\label{app:alternatives}
We considered three simpler rules. Each fixes one assignment by some criterion and then reports a different quantity. A system can then gain by changing the input to the rule instead of finding signals.

\begin{itemize}
\item \textbf{Maximize total credit, then report the ratio.} Adding a correct label can change which assignment wins and lower the reported ratio, so a system can gain by withholding a correct label. It is a defect of the definition, and a system can target it.
\item \textbf{Maximize credit only, ignoring evidence weight.} Removing a partial-credit relation can raise the score, which violates Proposition~\ref{prop:mono}.
\item \textbf{Count predictions in the denominator,} $2S/(|Q|+W)$, instead of matched weights. A perfect system then scores below 1 whenever some $p_g<1$, and dropping weakly supported reference labels from its output can \emph{raise} this score, while EW-F1 falls. This form rewards withholding weakly supported labels, which Proposition~\ref{prop:thresh} rules out for EW-F1.
\end{itemize}
EW-F1 removes the common cause: the score is the maximum of the quantity it reports. Propositions~\ref{prop:mono} and~\ref{prop:tie} are then theorems, not repairs.

\subsection{Worked example}
\label{app:worked}
Table~\ref{tab:ewf1-example} gives the earned credit $S(A)$ and the prediction mass $N'(A)$ behind the worked example of the main paper (Figure~\ref{fig:matching}). Its reference mass is $W=1.5$.

\begin{table}[t]
\centering
\begin{minipage}[b]{0.49\textwidth}
\captionsetup{position=top, skip=0pt}
\caption{Worked example of the main paper (Figure~\ref{fig:matching}). Reference: \emph{uncertainty} ($p_g{=}0.9$), \emph{hesitation} ($p_g{=}0.6$).}
\label{tab:ewf1-example}
\end{minipage}\hfill
\begin{minipage}[b]{0.49\textwidth}
\captionsetup{position=top, skip=0pt}
\caption{EW-F1 and plain cluster-weighted F1 on one clip. Reference: \texttt{Stress} ($p_g{=}0.90$), \texttt{Hesitation} ($p_g{=}0.25$).}
\label{tab:worked}
\end{minipage}

\vspace{-2pt}%
\begin{minipage}[t]{0.49\textwidth}
\centering
\vspace{0pt}%
\tablefont
\setlength{\tabcolsep}{4pt}
\begin{tabular}{@{}lrrrr@{}}
\toprule
\textbf{Prediction} & $S(A)$ & $N'(A)$ & \textbf{EW-F1} & \textbf{Exact F1} \\
\midrule
uncertainty, hesitation & 1.500 & 1.50 & 1.000 & 1.000 \\
confusion, hesitation & 1.194 & 1.50 & 0.796 & 0.500 \\
hesitation & 0.600 & 0.60 & 0.571 & 0.667 \\
boredom, hesitation & 0.600 & 1.60 & 0.387 & 0.500 \\
\bottomrule
\end{tabular}
\end{minipage}\hfill
\begin{minipage}[t]{0.49\textwidth}
\centering
\vspace{0pt}%
\tablefont
\setlength{\tabcolsep}{3pt}
\begin{tabular}{@{}lrr@{}}
\toprule
\textbf{Prediction} & \textbf{EW-F1} & \textbf{Plain} \\
\midrule
\texttt{Stress}, \texttt{Hesitation} & 1.000 & 1.000 \\
\texttt{Stress} & 0.878 & 0.667 \\
\texttt{Hesitation} & 0.357 & 0.667 \\
\texttt{Frustration}, \texttt{Hesitation} & 0.734 & 0.830 \\
\texttt{Stress}, \texttt{Hesitation}, \texttt{Engagement} & 0.697 & 0.800 \\
\bottomrule
\end{tabular}
\end{minipage}
\end{table}

A second clip has two reference signals. \texttt{Stress} has weight $0.90$, because three raters marked it with confidence. \texttt{Hesitation} has weight $0.25$, because one rater marked it and was unsure. Then $W=1.15$. Rows 2 and 3 of Table~\ref{tab:worked} show the purpose of the weight. A plain cluster-weighted F1 charges $0.333$ for each miss. EW-F1 charges $0.122$ for the weak miss and $0.643$ for the strong one.

\subsection{Error bars}
\label{app:errorbars}
Clips from one recording share speakers, settings, and raters, so they are not independent. We resample \emph{parent recordings} and keep every clip of a sampled recording together. The number of recordings, not clips, therefore sets the width of an interval. A maximum is not a smooth function of the data, and the ordinary bootstrap can be inconsistent for such a function~\citep{politis1994large}. We therefore use $m$-out-of-$n$ subsampling without replacement as the primary method (Algorithm~\ref{alg:interval}). An ordinary bootstrap interval is a secondary number only.

\begin{algorithm}[h]
\caption{95\% interval for one split.}
\label{alg:interval}
\small
\begin{algorithmic}[1]
\Require split with $n$ parent recordings; draws $B=2000$
\State $m\gets\lceil n^{0.7}\rceil$;\; $\theta_n\gets$ Algorithm~\ref{alg:dinkelbach} on the whole split
\For{$b\gets1$ \textbf{to} $B$}
  \State draw $m$ recordings without replacement, keeping all their clips
  \State $\theta_m^{(b)}\gets$ Algorithm~\ref{alg:dinkelbach} on that subset
  \State $t^{(b)}\gets\sqrt m\,(\theta_m^{(b)}-\theta_n)$
\EndFor
\State \Return $[\theta_n-t_{(0.975)}/\sqrt n,\ \theta_n-t_{(0.025)}/\sqrt n]$
\end{algorithmic}
\end{algorithm}

For paired comparisons, both systems are scored on the same draws and we report the interval of their difference. Two separate intervals can overlap while the paired difference is clear.

A per-signal interval needs reference support in at least 20 parent recordings; below that floor we report \textsc{not estimable}. The floor exists because a signal present in one recording is absent from a share near $e^{-1}\approx36.8\%$ of resamples, so empty cases dominate the distribution. The test partitions were built so that every signal reaches it (Supplement~\ref{app:partitions}).

Table~\ref{tab:width} gives the expected half-width of a per-signal interval as a function of the number of parent recordings. The values come from a simulation, not from the corpus: the simulated system has true per-signal score 0.75, each recording contributes two Bernoulli outcomes, and each row is the mean of 200 simulated datasets with a fixed seed. They indicate an order of magnitude. A half-width of $\pm0.05$ on one signal needs about 142 recordings, and $\pm0.03$ needs about 388.

\begin{table}[t]
\centering
\begin{minipage}[b]{0.36\textwidth}
\captionsetup{position=top, skip=0pt}
\caption{Simulated half-width of a 95\% per-signal interval, and the smallest detectable paired difference.}
\label{tab:width}
\end{minipage}\hfill
\begin{minipage}[b]{0.6\textwidth}
\captionsetup{position=top, skip=0pt}
\caption{Perturbation results on the 2{,}372-clip \sbvideo test reference (mean of 20 draws; SD at most 0.008), compared between EW-F1 and exact F1 (EW-F1/F1).}
\label{tab:supp-perturbation}
\end{minipage}

\vspace{-2pt}%
\begin{minipage}[t]{0.36\textwidth}
\centering
\vspace{0pt}%
\tablefont
\begin{tabular}{@{}rrr@{}}
\toprule
\textbf{Recordings} & \textbf{Half-width} & \textbf{Detectable diff.} \\
\midrule
10 & 0.172 & 0.100 \\
20 & 0.127 & 0.073 \\
40 & 0.092 & 0.053 \\
100 & 0.059 & 0.034 \\
196 & 0.043 & 0.024 \\
500 & 0.027 & 0.015 \\
\bottomrule
\end{tabular}
\end{minipage}\hfill
\begin{minipage}[t]{0.6\textwidth}
\centering
\vspace{0pt}%
\tablefont
\setlength{\tabcolsep}{3.5pt}
\begin{tabular}{@{}lccccc@{}}
\toprule
$r$ & Same & Adjacent & Unrelated & Delete & Insert \\
\midrule
0.1 & .979/.946 & .961/.942 & .933/.942 & .969/.969 & .964/.972 \\
0.2 & .959/.893 & .923/.883 & .864/.880 & .937/.937 & .935/.950 \\
0.3 & .938/.839 & .882/.823 & .796/.818 & .899/.898 & .906/.927 \\
0.4 & .922/.793 & .850/.774 & .739/.766 & .863/.863 & .884/.910 \\
0.5 & .903/.743 & .812/.716 & .673/.704 & .826/.825 & .866/.895 \\
\bottomrule
\end{tabular}
\end{minipage}
\end{table}

\subsection{The engagement head}
\label{app:engagement}
Every evaluation clip carries one engagement status: \texttt{disengaged}, \texttt{neutral}, or \texttt{engaged}. The status is a single choice, not a set, so EW-F1 has the wrong shape for it. We score it separately by unweighted macro-F1,
\begin{equation}
\mathrm{MF1}=\frac13\sum_{k}\frac{2\,\mathrm{TP}_k}{2\,\mathrm{TP}_k+\mathrm{FP}_k+\mathrm{FN}_k},
\label{eq:mf1}
\end{equation}
and a class with no true and no predicted positive scores 0. We never average MF1 with EW-F1. Accuracy is not the headline number. A constant predictor scores 0 on the two classes it never outputs, so its MF1 is at most $1/3$, whatever its accuracy. Every result also reports the full confusion matrix and the \emph{off-by-two rate}, the share of clips called \texttt{engaged} when the truth is \texttt{disengaged}, or the reverse. The classes are ordered, so that error is worse than a confusion with \texttt{neutral}, but MF1 charges both the same. Intervals use Algorithm~\ref{alg:interval} with Eq.~\ref{eq:mf1} as the score.

\subsection{Limits of the metric}
\label{app:limits}
\begin{enumerate}
\item \textbf{Retained labels only.} The merge drops crowd spans with no related expert span in the slice, and runs of one signal shorter than 2.0 seconds. EW-F1 therefore measures performance on labels that pass this gate, not robustness to all annotation uncertainty. The distribution of $p_g$ is truncated from below. That is a bias, not a variance, so no interval corrects it.
\item \textbf{Volume, not agreement.} The weight accumulates positive evidence and has no term for a rater who did not mark the signal.
\item \textbf{Optimistic alignment.} The score takes the best allowed assignment. It cannot invent credit, because $c$ and $p$ are fixed before scoring.
\item \textbf{Hand-set credit tiers.} The values $\alpha=0.66$ and $\beta=0.33$ encode the taxonomy. Leaderboard rankings are stable over a $3\times3$ grid (Supplement~\ref{sec:supp-leaderboard-metrics}), but a human-severity study has not yet calibrated the values.
\item \textbf{Rare classes.} The minority engagement class and most single signals are below the estimability floor on small sets. Per-signal intervals are wide (Table~\ref{tab:width}).
\end{enumerate}

\section{Metric Validation Details}
\label{sec:supp-validation}

All analyses in this section use \sbvideo; the same procedure applies to \sbaudio. Perturbations use the 2{,}372-clip \sbvideo test reference, which has 2{,}016 reference signals. Of these, 1{,}209 are eligible (see below); \emph{agreement}, \emph{confidence}, and \emph{interest} have no same-cluster alternative, so their labels stay unchanged in every setting. The credit-tier and evidence-weight analyses rescore the eight \sbvideo leaderboard systems on the same clips (Supplement~\ref{sec:supp-leaderboard-metrics}). All scores use the pinned release of the EW-F1 scorer. The analysis scripts and their full output are part of the release (\texttt{evaluation\_runs/metric\_validation/}).

A reference label is eligible for perturbation when it has same-cluster, adjacent, and unrelated alternatives, so all five perturbation types act on the same label population. Each eligible label is perturbed independently with probability $r$. A substitution draws its replacement uniformly from the alternatives of the requested relation that are not already present. Insertion adds the out-of-taxonomy label \emph{boredom}. Exact F1 is computed with the same matcher, with unit weights and $\alpha=\beta=0$.

Insertion behaves differently from the other perturbations. An inserted out-of-taxonomy label earns nothing and adds a full unit to $N'(A)$, whereas a deletion removes at most $p_g\le1$ from both credit and prediction mass. At low rates an insertion therefore costs more than a deletion, and the two curves cross between $r=0.2$ and $0.3$. At higher rates, insertion is the least damaging perturbation because the reference label is kept. This is the intended behavior: EW-F1 penalizes a spurious label but does not count it as a miss.

\label{sec:ewf1-validation}
The credit tiers still need a human check. We will sample prediction--truth confusions stratified by credit tier and show them to raters who do not see the metric's assignment. Raters grade the severity of each error on a fixed scale, or choose the worse of two confusions, and a rank correlation with a pre-registered acceptance threshold compares the human ordering with the ordering implied by the metric. This is the calibration that the main paper lists as pending (Section~\ref{sec:limitations}).

\section{Evaluation Harness}
\label{app:harness}
Every system on the leaderboard, including each frontier baseline, receives the same engineered harness rather than a generic instruction. It took several iterations of prompt design and output-contract hardening, and it carries much of the task knowledge that a trained model acquires from data. It has three parts: a signal catalog, decision rules, and an output contract.

The signal catalog is a system prompt of about 1{,}400 words that defines all twelve entries (ten scored signals plus engagement and disengagement). Each entry has a one-sentence definition and a list of observable cues, 70 cues in total, that name verbal, nonverbal, and prosodic evidence. The audio version of the catalog keeps only audio cues. A typical entry reads:
\begin{quote}\small
\textbf{Hesitation.} A temporary disruption in speaking or responding to a question. \emph{Observable cues:} the person delays or softens their answer; fills the gap with ``uh/um'' or restarts a phrase; looks away briefly before continuing; shows difficulty getting the words out; their last word stretches unnaturally long or drops into creaky voice.
\end{quote}

The decision rules separate the engagement decision from the signal list. They tell the model to focus on the primary speaker, to treat the clip as one unit, to use only observable evidence, and to omit weak, ambiguous, or merely possible labels; an empty signal list is a valid answer.

The output contract fixes the JSON schema, requires catalog names verbatim, and gives two worked examples, one with signals and one with none. The parser accepts case and whitespace variants of signal names. Each run records parse validity, so formatting failures are reported separately.

We also tested automatic prompt search with GEPA~\cite{agrawal2025gepa}, a reflective evolutionary prompt optimizer, through DSPy, on an earlier fine-tuned model. A reflection model read failed predictions and proposed edits to the instruction text around the signal catalog. The catalog and the output contract stayed fixed. Candidates were scored by signal-name F1 on 85 clips, 49 for search and 36 for selection. Most were hard cases: clips without audio, clips whose speaker is off camera, and clips with no reference signal. The best of nine candidates raised F1 on easy clips but did not make the model abstain when the speaker was off camera, where it usually predicted Hesitation from pauses or gaze shifts. We therefore used the search only to diagnose failure modes; the final rules were written and fixed by hand.

All systems receive the same clips and the same user turn. Gemini 4 Argon, Gemini 3.8 Flash, and Qwen3.5-Omni-Plus take the video with its audio track natively. GPT-6.1 Sol and Gemma-4-31B-it receive 16 frames sampled at 1\,fps with a transcript, and Muse Spark 1.3 receives the video with a transcript instead of the audio track. Outputs are scored by one evaluator from the unaggregated predictions.

The harness matters for the comparison. A frontier model given only the label names and a request for JSON must infer what each signal looks like, where the boundary between adjacent signals lies, and when to abstain. The harness answers these questions in the prompt. Frontier scores in the leaderboard therefore reflect the models under favorable conditions, and we expect scores under a minimal prompt to be lower. The gap between a trained model and a harnessed frontier model is thus a conservative estimate of the gap to frontier models as they are typically used.

\section{Models and Additional Results}
\label{app:results}
\subsection{Post-trained models}
\label{sec:supp-post-training}
Qwen2.5-Omni-7B is post-trained with supervised fine-tuning on the internal \sbvideo training partitions. Only language-side LoRA adapters are trained; the vision and audio encoders stay frozen, and training and evaluation use the same frame sampling (2~fps) and audio input. Targets follow the evaluation output contract, and about one fifth of training clips have no reference signal, so the model learns to abstain. For both post-trained models, the checkpoint is selected on the development partition, and the test set is scored once, after selection.

Pretraining alone does not solve the benchmark. On the development partitions, untuned Qwen2.5-Omni-7B scores 10.9 exact signal F1 and an untuned Gemma 4 E4B scores 20.3; after supervised training on the internal \sbvideo training partitions, the same backbones score 37.1 and 34.4. \sbaudio shows the same pattern: on its development partition, untuned Qwen2-Audio-7B-Instruct scores 17.8 exact signal F1, and the post-trained model 58.6.

The video and audio teacher models that produce pseudo-labels are scored against expert annotations on human-annotated YouTube clips, with the benchmark's EW-F1 scorer (Table~\ref{tab:pseudo-quality}). EW-F1 intervals use Algorithm~\ref{alg:interval} (recording-level subsampling without replacement, $m=\lceil n^{0.7}\rceil$, 2,000 draws).

\begin{table}[h]
\centering
\caption{Pseudo-label quality of the teacher models against expert annotations.}
\label{tab:pseudo-quality}
\tablefont
\begin{tabular*}{\textwidth}{@{\extracolsep{\fill}}lrrrr@{}}
\toprule
\textbf{Teacher} & \textbf{Clips} & \textbf{Recordings} & \textbf{EW-F1} & \textbf{95\% CI} \\
\midrule
Video & 758 & 62  & 54.8 & [50.4, 61.0] \\
Audio & 604 & 106 & 65.4 & [62.2, 69.1] \\
\bottomrule
\end{tabular*}
\end{table}

\subsection{Leaderboard details}
Table~\ref{tab:frontier} gives the scores behind main-paper Figure~\ref{fig:leaderboard}.

\begin{table}[h]
\centering
\caption{Social Signal Bench leaderboard (main paper Figure~\ref{fig:leaderboard}) on the 2{,}372-clip test set of \sbvideo (85 recordings) and the held-out test set of \sbaudio under the shared harness and evaluator. Systems are ranked by EW-F1 within each dataset; \emph{Eng.\ MF1} is engagement macro-F1. The datasets use different signal subsets, so scores should only be compared within each dataset.}
\label{tab:frontier}
\tablefont
\setlength{\tabcolsep}{4pt}
\begin{tabular*}{\textwidth}{@{\extracolsep{\fill}}llcc@{}}
\toprule
\textbf{System} & \textbf{Access}
& \textbf{EW-F1 $\uparrow$} & \textbf{Eng.\ MF1 $\uparrow$} \\
\midrule
\multicolumn{4}{@{}l}{\textbf{\sbvideo}} \\
\midrule
Muse Spark 1.3 & API & 40.9 & 43.0 \\
Gemini 3.8 Flash & API & 40.6 & 35.1 \\
Qwen3.5-Omni-Plus & open & 40.6 & 40.0 \\
Gemini 4 Argon & API & 40.4 & 39.4 \\
GPT-6.1 Sol & API & 39.6 & 33.5 \\
Gemma-4-31B-it & open & 35.3 & 38.2 \\
Qwen2.5-Omni-7B (untuned) & open & 20.6 & 20.7 \\
\midrule
\textbf{Qwen2.5-Omni-7B (post-trained)}
& open & \textbf{44.0} & \textbf{58.4} \\
\midrule
\multicolumn{4}{@{}l}{\textbf{\sbaudio}} \\
\midrule
Gemini 3.1 Pro & API & 54.0 & 46.5 \\
Gemini 3.8 Flash & API & 51.0 & 35.8 \\
Qwen3.5-Omni-Plus & open & 47.5 & 47.2 \\
GPT-Audio & API & 44.4 & 37.7 \\
Kimi-Audio-7B-Instruct & open & 29.8 & 26.2 \\
Qwen2-Audio-7B-Instruct & open & 28.7 & 12.4 \\
Audio-Flamingo-3 & open & 25.4 & 32.9 \\
Ultravox-v0.5-Llama-3.1-8B & open & 23.6 & 31.1 \\
\midrule
\textbf{Qwen2-Audio-7B-Instruct (post-trained)}
& open & \textbf{58.1} & 45.8 \\
\bottomrule
\end{tabular*}
\end{table}

\label{sec:supp-leaderboard-metrics}
Table~\ref{tab:supp-leaderboard-metrics} rescores the eight \sbvideo systems of Table~\ref{tab:frontier} on the same 2{,}372 clips with exact micro- and macro-F1 over the ten signals and with uniform evidence weights ($p_g=1$). The post-trained model ranks first under EW-F1, exact micro-F1, and uniform weights, and third under exact macro-F1, behind Qwen3.5-Omni-Plus and Gemini 4 Argon. Kendall's $\tau$ between the EW-F1 ranking and the exact micro-F1, exact macro-F1, and uniform-weight rankings is 0.71, 0.57, and 0.57. Every rank change involves Muse Spark 1.3, Gemini 3.8 Flash, Qwen3.5-Omni-Plus, and Gemini 4 Argon, whose EW-F1 scores span 40.4 to 40.9, apart from the exact macro-F1 placement of the post-trained model. On the $3\times3$ grid of credit tiers $\alpha\in\{0.5,0.66,0.8\}\times\beta\in\{0.17,0.33,0.5\}$, the post-trained model ranks first and GPT-6.1 Sol, Gemma-4-31B-it, and the untuned backbone rank sixth to eighth in every cell; only the same four frontier models swap places. The largest change of any score is 3.1 points (untuned backbone at $\alpha=0.8$, $\beta=0.5$), and at most 1.9 points for any other system. Exact matches supply 89.2\% of the post-trained model's earned credit, same-cluster matches 4.7\%, and adjacent-cluster matches 6.1\%. Paired differences, post-trained model minus baseline (2{,}000 draws over the 85 recordings, seed 42), are $+3.2$ [$+0.8$, $+5.9$] for Muse Spark 1.3, $+3.5$ [$+1.0$, $+6.9$] for Gemini 3.8 Flash, $+3.5$ [$+1.2$, $+6.7$] for Qwen3.5-Omni-Plus, $+3.7$ [$+1.4$, $+7.0$] for Gemini 4 Argon, $+4.4$ [$+2.2$, $+7.4$] for GPT-6.1 Sol, and $+8.8$ [$+6.3$, $+12.7$] for Gemma-4-31B-it; every interval excludes zero. Muse Spark 1.3 returned 20 empty responses, which are scored as predicting no signal; all other outputs parse.

\begin{table}[h]
\centering
\caption{The Social Signal Bench Video leaderboard (2{,}372 clips) under EW-F1, exact micro- and macro-F1, and EW-F1 with uniform evidence weights. Ranks in parentheses.}
\label{tab:supp-leaderboard-metrics}
\tablefont
\setlength{\tabcolsep}{3pt}
\begin{tabular*}{\textwidth}{@{\extracolsep{\fill}}lcccc@{}}
\toprule
\textbf{System} & \textbf{EW-F1} & \textbf{Micro-F1} & \textbf{Macro-F1} & \textbf{Uniform} \\
\midrule
Qwen2.5-Omni-7B (post-trained) & 44.0 (1) & 39.2 (1) & 31.0 (3) & 44.0 (1) \\
Muse Spark 1.3 & 40.9 (2) & 35.6 (4) & 29.8 (4) & 40.9 (5) \\
Gemini 3.8 Flash & 40.6 (3) & 35.2 (5) & 29.4 (5) & 41.2 (4) \\
Qwen3.5-Omni-Plus & 40.6 (4) & 38.6 (2) & 33.1 (1) & 41.9 (3) \\
Gemini 4 Argon & 40.4 (5) & 37.1 (3) & 32.8 (2) & 42.3 (2) \\
GPT-6.1 Sol & 39.6 (6) & 34.7 (6) & 28.7 (6) & 40.5 (6) \\
Gemma-4-31B-it & 35.3 (7) & 33.0 (7) & 26.8 (7) & 37.4 (7) \\
Qwen2.5-Omni-7B (untuned) & 20.6 (8) & 13.7 (8) & 10.2 (8) & 21.1 (8) \\
\bottomrule
\end{tabular*}
\end{table}

On \sbaudio (Table~\ref{tab:supp-leaderboard-metrics-audio}), the post-trained model ranks first under every metric. Uniform weights raise every score by 0.7 to 2.0 points and change no rank ($\tau=1$); Kendall's $\tau$ against the exact micro- and macro-F1 rankings is 0.94 and 0.89, and every rank change involves the three lowest-scoring open models.
\begin{table}[h]
\centering
\caption{The Social Signal Bench Audio leaderboard (1{,}094 clips) under EW-F1, exact micro- and macro-F1 over the six audio signals, and EW-F1 with uniform evidence weights.}
\label{tab:supp-leaderboard-metrics-audio}
\tablefont
\setlength{\tabcolsep}{3pt}
\begin{tabular*}{\textwidth}{@{\extracolsep{\fill}}lcccc@{}}
\toprule
\textbf{System} & \textbf{EW-F1} & \textbf{Micro-F1} & \textbf{Macro-F1} & \textbf{Uniform} \\
\midrule
Qwen2-Audio-7B-Instruct (post-trained) & 58.1 (1) & 54.7 (1) & 47.4 (1) & 58.8 (1) \\
Gemini 3.1 Pro & 54.0 (2) & 51.2 (2) & 44.7 (2) & 55.3 (2) \\
Gemini 3.8 Flash & 51.0 (3) & 47.1 (3) & 43.2 (3) & 52.4 (3) \\
Qwen3.5-Omni-Plus & 47.5 (4) & 44.1 (4) & 42.0 (4) & 49.0 (4) \\
GPT-Audio & 44.4 (5) & 42.3 (5) & 37.8 (5) & 45.8 (5) \\
Kimi-Audio-7B-Instruct & 29.8 (6) & 26.7 (6) & 20.8 (6) & 31.2 (6) \\
Qwen2-Audio-7B-Instruct (untuned) & 28.7 (7) & 20.7 (8) & 17.8 (8) & 30.0 (7) \\
Audio-Flamingo-3 & 25.4 (8) & 25.4 (7) & 14.2 (9) & 27.5 (8) \\
Ultravox-v0.5 (Llama-3.1-8B) & 23.6 (9) & 16.4 (9) & 20.7 (7) & 24.7 (9) \\
\bottomrule
\end{tabular*}
\end{table}

By domain, results stay close to the overall score. On the 2{,}372 test clips (85 groups), post-trained Qwen2.5-Omni-7B scores 44.0 EW-F1 [41.0, 47.0] (2{,}000 recording-level draws, seed 42), and it scores 47.5 [44.9, 50.6] on dyadic conversations (4 groups), 43.7 [39.9, 47.8] on online video (75 groups), and 40.8 [38.3, 43.0] on sales calls (6 groups); the dyadic and sales intervals rest on few groups.

\subsection{Modality ablation and probes}\label{sec:supp-modality}
Figure~\ref{fig:supp-modality} evaluates the post-trained checkpoint on all 2{,}372 test clips with one stream removed, using the leaderboard inference settings. \emph{Video only} keeps the video stream unchanged and replaces the audio with a silent track; \emph{audio only} keeps the audio and replaces every frame with a black frame of the same size, rate and duration. All outputs parse. With black frames the model abstains on most clips (92\% of predictions contain no signal), so the audio-only score reflects the trained model's behavior without visual input rather than an upper bound on what the audio carries.

\begin{figure}[t]
\centering
\begin{minipage}[b]{0.48\textwidth}\centering
\resizebox{\linewidth}{!}{\begin{tikzpicture}[
    font=\scriptsize,
    full/.style={circle, fill=cvprblue, inner sep=0pt, minimum size=5.5pt},
    abl/.style={circle, fill=black!65, inner sep=0pt, minimum size=4.6pt},
    dom/.style={circle, draw=black!55, fill=white, line width=0.45pt, inner sep=0pt, minimum size=3pt},
    stem/.style={black!20, line width=0.5pt},
    ci/.style={line width=0.9pt, line cap=round},
    val/.style={anchor=west, inner sep=1pt},
]
  \def\xa{0}    \def\sa{0.068}   %
  \def\xb{4.05} \def\sb{0.03}    %
  \def\dy{1.6}   %
  \foreach \t in {0,10,...,50} {
    \draw[black!8] (\xa+\sa*\t,0.25) -- (\xa+\sa*\t,-2*\dy-0.3);
    \node[below, text=black!60] at (\xa+\sa*\t,-2*\dy-0.3) {\t};
  }
  \foreach \t in {0,20,...,60} {
    \draw[black!8] (\xb+\sb*\t,0.25) -- (\xb+\sb*\t,-2*\dy-0.3);
    \node[below, text=black!60] at (\xb+\sb*\t,-2*\dy-0.3) {\t};
  }
  \node[anchor=south, font=\scriptsize\bfseries] at (\xa+\sa*25,0.3) {EW-F1};
  \node[anchor=south, font=\scriptsize\bfseries] at (\xb+\sb*35,0.3) {Engagement MF1};
  \def\row#1#2#3#4#5#6#7#8#9{%
    \pgfmathsetmacro{\yy}{-(#1)*\dy}
    \node[anchor=east, align=right, inner sep=1pt] at (-0.12,\yy) {#2\\[-1pt]{\tiny\color{black!55}#3}};
    \draw[stem] (\xa,\yy) -- (\xa+\sa*#5,\yy);
    \foreach \d in {#8} \node[dom] at (\xa+\sa*\d,\yy-0.17) {};
    \draw[ci, #4] (\xa+\sa*#6,\yy) -- (\xa+\sa*#7,\yy);
    \node[#4] (p) at (\xa+\sa*#5,\yy) {};
    \node[val, font=\scriptsize\bfseries] at (\xa+\sa*#7+0.03,\yy+0.01) {#5};
    \draw[stem] (\xb,\yy) -- (\xb+\sb*#9,\yy);
    \node[#4] at (\xb+\sb*#9,\yy) {};
    \node[val] at (\xb+\sb*#9+0.08,\yy) {#9};
  }
  \tikzset{fullc/.style={full, cvprblue}, ablc/.style={abl, black!65}}
  \row{0}{Audio + video}{full input}{fullc}{44.0}{41.0}{47.0}{45.5,39.8,47.5,40.8}{58.4}
  \row{1}{Video only}{silent audio}{ablc}{22.1}{19.0}{25.5}{22.3,19.6,20.2,24.4}{63.5}
  \row{2}{Audio only}{black frames}{ablc}{15.4}{11.1}{19.9}{11.3,16.7,25.1,9.3}{14.1}
  \node[dom] at (\xa+0.12,-2*\dy-0.95) {};
  \node[anchor=west, text=black!60] at (\xa+0.2,-2*\dy-0.95) {one domain};
  \draw[ci, black!65] (\xa+1.75,-2*\dy-0.95) -- ++(0.3,0);
  \node[anchor=west, text=black!60] at (\xa+2.1,-2*\dy-0.95) {95\% interval};
\end{tikzpicture}}
\end{minipage}\hfill
\begin{minipage}[b]{0.48\textwidth}\centering
\resizebox{\linewidth}{!}{\begin{tikzpicture}[
    font=\scriptsize,
    pt/.style={circle, fill=cvprblue, inner sep=0pt, minimum size=4.6pt},
    ptx/.style={circle, draw=cvprblue, fill=white, line width=0.7pt, inner sep=0pt, minimum size=4.6pt},
    em/.style={rectangle, fill=black!60, inner sep=0pt, minimum size=4pt},
    ci/.style={line width=0.8pt, line cap=round},
    cem/.style={black!60}, cpt/.style={cvprblue}, cptx/.style={cvprblue},
]
  \def\ys{0.062} \def\yl{20} \def\dx{0.62}
  \def\panel#1#2#3{%
    \foreach \t in {20,30,...,60} \draw[black!8] (#1-0.2,{(\t-\yl)*\ys}) -- (#1+#3*\dx+0.05,{(\t-\yl)*\ys});
    \node[anchor=south, font=\scriptsize\bfseries] at (#1+#3*\dx/2-0.08,{(62-\yl)*\ys}) {#2};
    \draw[black!30, densely dashed] (#1+0.62*\dx,-0.05) -- (#1+0.62*\dx,{(60-\yl)*\ys});
  }
  \def\cat#1#2#3#4#5#6#7#8{%
    \pgfmathsetmacro{\xx}{#1+(#2)*\dx}
    \draw[ci, c#6] (\xx,{(#4-\yl)*\ys}) -- (\xx,{(#5-\yl)*\ys});
    \node[#6] at (\xx,{(#3-\yl)*\ys}) {};
    \node[below, inner sep=1pt] at (\xx,-0.05) {#7};
    \node[below, inner sep=1pt, text=black!50, font=\tiny] at (\xx,-0.32) {#8};
  }
  \foreach \t in {20,30,...,60} \node[left, text=black!60, inner sep=1pt] at (-0.2,{(\t-\yl)*\ys}) {\t};
  \node[rotate=90, anchor=south] at (-0.62,{20*\ys}) {score (\%)};
  \node[anchor=east, text=black!50, font=\tiny] at (-0.22,-0.45) {clips};

  \panel{0.1}{Audiovisual}{5}
  \cat{0.1}{0.2}{47.4}{41.1}{54.2}{em}{0}{81}
  \cat{0.1}{1}{49.3}{43.2}{55.3}{pt}{1}{97}
  \cat{0.1}{2}{42.9}{37.6}{47.7}{pt}{2}{77}
  \cat{0.1}{3}{40.9}{31.9}{52.6}{ptx}{3--4}{47}
  \cat{0.1}{4}{32.5}{22.5}{47.3}{ptx}{5+}{11}

  \panel{3.65}{Audio only}{4}
  \cat{3.65}{0.2}{32.2}{27.3}{37.5}{em}{0}{156}
  \cat{3.65}{1}{56.4}{50.7}{62.1}{pt}{1}{132}
  \cat{3.65}{2}{47.8}{38.9}{58.1}{pt}{2}{79}
  \cat{3.65}{3}{48.1}{44.2}{54.1}{ptx}{3+}{24}

  \node[below=9pt] at (3.4,-0.32) {reference signals per clip};
  \begin{scope}[shift={(0.25,-1.25)}]
    \node[em] at (0,0) {}; \node[anchor=west] at (0.08,0) {empty-selection accuracy};
    \node[pt] at (3.0,0) {}; \node[anchor=west] at (3.08,0) {EW-F1};
    \node[ptx] at (4.1,0) {}; \node[anchor=west] at (4.18,0) {${<}20$ parents};
  \end{scope}
\end{tikzpicture}}
\end{minipage}

\medskip
\begin{minipage}[t]{0.48\textwidth}
\caption{\textbf{Modality ablation} of post-trained Qwen2.5-Omni-7B on the 2{,}372 \sbvideo test clips. Dots are EW-F1 with 95\% recording-level intervals (85 recordings); hollow circles are the four domains (online video A and B, dyadic conversations, sales calls). Removing either stream halves EW-F1 or worse. Engagement survives silent audio but not black frames.}
\label{fig:supp-modality}
\end{minipage}\hfill
\begin{minipage}[t]{0.48\textwidth}
\caption{\textbf{Human performance by reference density.} Mean of the individual trained raters on shared test clips (five raters on audiovisual clips, four on audio-only clips), with 95\% parent-grouped bootstrap intervals (2{,}000 draws). Clips without a reference signal are scored by correct empty selection, not EW-F1 (left of the dashed line). Hollow markers rest on fewer than 20 parent recordings and are exploratory.}
\label{fig:human-density}
\end{minipage}
\end{figure}

Frozen linear probes show where signal information is readable inside the trained models. Probe training, selection, and test use different partitions. On the 2{,}372 audiovisual test clips, macro AP over the ten signals is 17.0 (Qwen) and 17.7 (Gemma) from pooled audio features, 15.3 and 14.1 from concatenated audio and video, and 24.2 and 23.6 from the decoder state at the last prompt token. The decoder exceeds the best unimodal readout by $+7.2$ $[4.2,10.0]$ and $+5.8$ $[3.5,8.7]$ points (500 paired recording-level bootstrap draws). Parameter-matched nonlinear heads on pooled audio exceed those on pooled video by 5.3 and 4.0 points. Pooling can discard brief visual cues, so this result bounds what pooled features expose. It does not show that vision is uninformative.

The audio-only test clips show a similar pattern. Macro AP over the six signals is 44.3 from pooled audio-encoder features and 48.4 from pooled decoder states at audio-token positions, against 53.8 from the decoder state at the last prompt token. The last-token readout exceeds the best pooled readout by $+5.4$ $[1.3,9.5]$ points (500 paired recording-level bootstrap draws). Parameter-matched nonlinear heads (about 1M parameters) give the same ordering, with 47.7 for the last-token readout and 44.3 for pooled decoder states, but the difference of $+3.3$ $[-0.7,6.2]$ includes zero. Within pooled readouts, decoder states exceed encoder features by $+4.2$ $[-0.6,8.6]$ points with linear heads and $+5.3$ $[0.9,12.4]$ with nonlinear heads. Pooling averages over the clip and can discard brief acoustic cues, so this result bounds what pooled features expose. It does not show that the pooled encoder carries no signal information.

\section{Human Reference Performance}
\label{app:human-reference}
\subsection{Audiovisual raters}
Five annotators provided 1,937 deduplicated judgments on 452 test clips, and all five labeled the same 313 clips from 36 parent videos. We score each person's selections against the frozen test reference with canonical EW-F1 and average the five corpus scores with equal weight. An omitted signal field means that no signal was selected; a missing annotation row is not imputed. The mean on the shared clips is 38.9 EW-F1, with a 95\% interval of [34.9, 42.6] from 2,000 draws of joint parent-video-grouped subsampling. This estimates individual performance; it is not a score for pooled human predictions.

Figure~\ref{fig:human-density} (left) stratifies the shared clips by the number of reference social signals, excluding engagement. Within each stratum, intervals use 2,000 joint parent-video percentile bootstrap draws (seed 42); they keep all rater judgments and score repeated clip copies separately, so they differ from the grouped-subsampling interval of the overall estimate. The intervals condition on the five observed raters, the frozen reference, and the observed coverage, and do not describe uncertainty over a wider annotator population.

Clips without a reference signal need a separate score. On a corpus whose references are all empty, the scorer returns EW-F1 of 100 only if every prediction is empty, and 0 otherwise. Each annotator predicts at least one signal on these 81 clips, so each scores 0, which hides partial success. We therefore report correct empty-selection accuracy: its mean is 47.4\% [41.1\%, 54.2\%], the complement of the 52.6\% [45.8\%, 58.9\%] rate of predicting any signal. Overall EW-F1 includes empty-reference clips. There, unmatched predictions add prediction cost without earning credit, while correct empty selections add neither credit nor cost. Overall EW-F1 is thus sensitive to background false positives but does not measure empty-class accuracy.

Leaving out one parent video at a time gives mean EW-F1 ranges of 38.4--45.2 for 3--4 signals and 28.1--36.4 for 5+ signals, with maximum absolute shifts of 4.3 and 4.4 points. The decline with density is descriptive: no paired or multiplicity-adjusted trend test establishes a difficulty ordering. Density is not annotatability, and clips with empty references are not necessarily unsuitable. Three further factors limit interpretation: complete-case selection, the difference between full-video reference construction and clip-level tagging, and the unestablished independence between reference creators and these raters. A test-selected, video-blind constant prediction scores 39.1 on the same clips. It is an exploratory control for the metric scale, not a held-out baseline, and individual raters do not beat this prior tuned to the subset. The results therefore set neither a human ceiling nor a ranking against models. Such a comparison needs aligned predictions on the same clips with paired parent-video uncertainty, and development-set model scores cannot be compared with this test-subset estimate. Stratum scores also cannot be clip-weighted to recover the overall corpus EW-F1.

\subsection{Audio raters}
Five trained raters labeled the audio track of clips from the \sbvideo test partition, hearing only the clip audio and choosing among the six audio signals and engagement status. We score them against the frozen test reference for these clips, which was annotated with video visible. Following the audio protocol, the four non-audio signals are removed from the reference and clips whose reference held only those signals are dropped. Skipped tasks are not imputed. One rater stopped using the signal field after 46 clips and is excluded. The remaining four share 391 clips from 40 parent recordings. Their mean EW-F1 is 40.6, with a 95\% joint parent-recording grouped-subsampling interval of [36.8, 44.5] from 2,000 draws (individual raters 30.7--44.5). Figure~\ref{fig:human-density} (right) stratifies these clips in the same way; only one clip carries five or more audio signals, so strata of three or more are merged. Correct empty selection varies widely across raters (1.3\%--83.3\%), so the zero-signal point mainly reflects how readily each rater marks a signal. Leaving out one parent recording at a time gives 46.8--50.1 for the 3+ stratum (maximum shift 2.0 points). The best test-selected constant rule, \emph{confidence} and \emph{hesitation} on every clip, scores 36.2. The rater mean exceeds it by 4.4 [$-0.2$, $+8.1$] (joint grouped subsampling, not adjusted for the rule search), so audio raters also do not clearly beat a prior tuned to this subset.

{\small
\bibliographystyle{ieeenat_fullname}
\bibliography{references}
}

\end{document}